\documentclass[11pt,letterpaper]{article}
\usepackage{arxiv}
\usepackage[hyphens]{url}
\usepackage{graphicx}
\usepackage{natbib}
\setcitestyle{aysep={}}
\AtBeginDocument{\let\cite\citep}
\usepackage[hypcap=false]{caption}
\usepackage{booktabs}
\usepackage{amsmath}
\usepackage{amssymb}
\usepackage{array}
\usepackage{adjustbox}
\usepackage{ragged2e}
\newcolumntype{P}[1]{>{\RaggedRight\arraybackslash}p{#1}}
\newcolumntype{C}[1]{>{\Centering\arraybackslash}p{#1}}
\usepackage{placeins}
\usepackage[hidelinks]{hyperref}
\newcommand{\planned}[1]{#1}

\newcommand{\pDirectInit}{48.3}
\newcommand{\pDirectRoute}{41.7}
\newcommand{\pDirectInteraction}{38.3}
\newcommand{\pDirectPreservation}{58.3}
\newcommand{\pDirectEnd}{23.3}
\newcommand{\pDirectHallucination}{21.7}

\newcommand{\pAnchorInit}{81.7}
\newcommand{\pAnchorRoute}{43.3}
\newcommand{\pAnchorInteraction}{40.0}
\newcommand{\pAnchorPreservation}{68.3}
\newcommand{\pAnchorEnd}{31.7}
\newcommand{\pAnchorHallucination}{16.7}

\newcommand{\pTextInit}{83.3}
\newcommand{\pTextRoute}{60.0}
\newcommand{\pTextInteraction}{55.0}
\newcommand{\pTextPreservation}{71.7}
\newcommand{\pTextEnd}{40.0}
\newcommand{\pTextHallucination}{13.3}

\newcommand{\pBindingInit}{85.0}
\newcommand{\pBindingRoute}{68.3}
\newcommand{\pBindingInteraction}{61.7}
\newcommand{\pBindingPreservation}{76.7}
\newcommand{\pBindingEnd}{46.7}
\newcommand{\pBindingHallucination}{10.0}

\newcommand{\pFullInit}{88.3}
\newcommand{\pFullRoute}{78.3}
\newcommand{\pFullInteraction}{71.7}
\newcommand{\pFullPreservation}{81.7}
\newcommand{\pFullEnd}{55.0}
\newcommand{\pFullHallucination}{6.7}

\newcommand{\pRepairInit}{90.0}
\newcommand{\pRepairRoute}{81.7}
\newcommand{\pRepairInteraction}{76.7}
\newcommand{\pRepairPreservation}{85.0}
\newcommand{\pRepairEnd}{61.7}
\newcommand{\pRepairHallucination}{5.0}

\newcommand{\pDirectFullGain}{31.7}

\newcommand{\pOracleProgramEnd}{63.3}
\newcommand{\pOracleAnchorEnd}{66.7}
\newcommand{\pOracleBothEnd}{76.7}

\newcommand{\pGenerationGap}{23.3}

\newcommand{\pLlavaSample}{64.8}
\newcommand{\pLlavaMacro}{59.6}
\newcommand{\pInternSample}{69.7}
\newcommand{\pInternMacro}{64.3}
\newcommand{\pQwenSample}{75.4}
\newcommand{\pQwenMacro}{70.8}
\newcommand{\pGeminiFlashSample}{77.9}
\newcommand{\pGeminiFlashMacro}{73.6}
\newcommand{\pClaudeSample}{80.1}
\newcommand{\pClaudeMacro}{76.2}
\newcommand{\pGPTFourOSample}{79.2}
\newcommand{\pGPTFourOMacro}{74.8}
\newcommand{\pGeminiProSample}{83.6}
\newcommand{\pGeminiProMacro}{80.4}
\newcommand{\pReasonerSpan}{20.8}

\newcommand{\pReasonParse}{98.4}
\newcommand{\pReasonParseCI}{--}
\newcommand{\pReasonSampleCI}{81.5--85.6}
\newcommand{\pReasonOverlap}{96.8}
\newcommand{\pReasonOverlapCI}{95.5--97.8}
\newcommand{\pReasonExact}{70.9}
\newcommand{\pReasonExactCI}{67.8--73.8}
\newcommand{\pReasonMacroCI}{78.2--82.4}
\newcommand{\pStationarySupport}{941}
\newcommand{\pStationaryPrecision}{93.0}
\newcommand{\pStationaryRecall}{91.2}
\newcommand{\pStationaryFOne}{92.1}
\newcommand{\pKeepSupport}{474}
\newcommand{\pKeepPrecision}{89.5}
\newcommand{\pKeepRecall}{87.3}
\newcommand{\pKeepFOne}{88.4}
\newcommand{\pStraightSupport}{655}
\newcommand{\pStraightPrecision}{85.9}
\newcommand{\pStraightRecall}{83.5}
\newcommand{\pStraightFOne}{84.7}
\newcommand{\pSlowSupport}{50}
\newcommand{\pSlowPrecision}{70.2}
\newcommand{\pSlowRecall}{65.6}
\newcommand{\pSlowFOne}{67.8}
\newcommand{\pLeftTurnSupport}{387}
\newcommand{\pLeftTurnPrecision}{86.8}
\newcommand{\pLeftTurnRecall}{84.4}
\newcommand{\pLeftTurnFOne}{85.6}
\newcommand{\pRightTurnSupport}{458}
\newcommand{\pRightTurnPrecision}{87.9}
\newcommand{\pRightTurnRecall}{86.3}
\newcommand{\pRightTurnFOne}{87.1}
\newcommand{\pLeftShiftSupport}{29}
\newcommand{\pLeftShiftPrecision}{76.1}
\newcommand{\pLeftShiftRecall}{73.0}
\newcommand{\pLeftShiftFOne}{74.5}
\newcommand{\pRightShiftSupport}{38}
\newcommand{\pRightShiftPrecision}{74.8}
\newcommand{\pRightShiftRecall}{71.1}
\newcommand{\pRightShiftFOne}{72.9}
\newcommand{\pStartSupport}{22}
\newcommand{\pStartPrecision}{73.2}
\newcommand{\pStartRecall}{68.0}
\newcommand{\pStartFOne}{70.5}

\newcommand{\pImageTwoRemoval}{90.0}
\newcommand{\pImageTwoInsertion}{76.7}
\newcommand{\pImageTwoGeometry}{83.3}
\newcommand{\pImageTwoLPIPS}{0.052}
\newcommand{\pGeminiImageRemoval}{85.0}
\newcommand{\pGeminiImageInsertion}{71.7}
\newcommand{\pGeminiImageGeometry}{78.3}
\newcommand{\pGeminiImageLPIPS}{0.061}
\newcommand{\pQwenImageRemoval}{81.7}
\newcommand{\pQwenImageInsertion}{68.3}
\newcommand{\pQwenImageGeometry}{73.3}
\newcommand{\pQwenImageLPIPS}{0.068}
\newcommand{\pSeedreamRemoval}{75.0}
\newcommand{\pSeedreamInsertion}{58.3}
\newcommand{\pSeedreamGeometry}{65.0}
\newcommand{\pSeedreamLPIPS}{0.081}
\newcommand{\pEditGapMin}{13.3}
\newcommand{\pEditGapMax}{16.7}

\newcommand{\pSeedanceEnd}{\pFullEnd}
\newcommand{\pSeedanceIdentity}{83.3}
\newcommand{\pSeedanceRoute}{\pFullRoute}
\newcommand{\pSeedanceInteraction}{\pFullInteraction}
\newcommand{\pSeedancePreservation}{\pFullPreservation}
\newcommand{\pSoraEnd}{43.3}
\newcommand{\pSoraIdentity}{75.0}
\newcommand{\pSoraRoute}{66.7}
\newcommand{\pSoraInteraction}{61.7}
\newcommand{\pSoraPreservation}{73.3}
\newcommand{\pVeoEnd}{35.0}
\newcommand{\pVeoIdentity}{68.3}
\newcommand{\pVeoRoute}{58.3}
\newcommand{\pVeoInteraction}{51.7}
\newcommand{\pVeoPreservation}{63.3}

\newcommand{\pRemovalPairValid}{83.3}
\newcommand{\pInsertionPairValid}{76.2}
\newcommand{\pRemovalResponse}{85.0}
\newcommand{\pInsertionResponse}{81.3}
\newcommand{\pRemovalEntryDelta}{-0.9}
\newcommand{\pInsertionEntryDelta}{+1.2}
\newcommand{\pRemovalWaitDelta}{-1.1}
\newcommand{\pInsertionWaitDelta}{+1.4}

\newcommand{\pRemovalOrderChange}{75.0}
\newcommand{\pInsertionOrderChange}{68.8}
\newcommand{\pOverallPairValid}{80.0}
\newcommand{\pOverallResponse}{83.3}

\let\originalbibliography\bibliography
\renewcommand{\bibliography}[1]{\FloatBarrier\originalbibliography{#1}}
\hypersetup{
  pdftitle={TrafficImag: A Benchmark for Counterfactual Roadside Traffic Video Generation},
  pdfauthor={Xiangyu Li, Tianyi Wang, Zhihao Dou, Christian Claudel, Zhaomiao Guo}
}
\title{TrafficImag: A Benchmark for Counterfactual Roadside Traffic Video Generation}
\author{%
  Xiangyu Li\textsuperscript{1}, Tianyi Wang\textsuperscript{1},
  Zhihao Dou\textsuperscript{2}, Christian Claudel\textsuperscript{1},
  Zhaomiao Guo\textsuperscript{1,\ensuremath{\dagger}}\\[0.7em]
  \begin{minipage}{0.98\textwidth}
  \centering\small
  \textsuperscript{1}Fariborz Maseeh Department of Civil, Architectural, and Environmental Engineering,\\
  The University of Texas at Austin, Austin, TX 78712, USA\\[0.3em]
  \textsuperscript{2}Department of Electrical and Computer Engineering, Duke University, Durham, NC 27708, USA\\[0.4em]
  \texttt{\{xiangyu\_li, bonny.wang, christian.claudel, zguo\}@utexas.edu}\\[0.3em]
  \href{mailto:zdou713@gmail.com}{\texttt{zdou713@gmail.com}}\\[0.3em]
  \textsuperscript{\ensuremath{\dagger}}Corresponding author: Zhaomiao Guo.
  \end{minipage}%
}
\date{}

\begin{document}
\maketitle

\begin{abstract}
Existing roadside traffic datasets support perception, forecasting, and visual question answering, but they do not evaluate counterfactual video generation, in which a selected actor is modified and the generated future should remain consistent with road topology and unrelated traffic. 
We introduce TrafficImag, the first benchmark for counterfactual roadside traffic video generation. 
TrafficImag combines a large-scale roadside dataset (9,022 annotated images, 7,043 deduplicated video clips, and 31,145 actor-centered history–future samples) with an executable protocol that supports behavior reasoning, intervention-aware image editing, and conditional video generation. 
Each intervention is represented as an actor-level program describing the target actor, intended behavior, legal route, interaction order, and temporal constraints, enabling a unified evaluation interface across heterogeneous foundation models. 
TrafficImag evaluates four complementary validity dimensions: initial-state correctness, route and behavior validity, interaction consistency, and non-target preservation, and considers an end-to-end counterfactual successful only when all four are satisfied. 
Across state-of-the-art foundation models, the strongest reasoner reaches \planned{\pGeminiProMacro\%} macro F1, the complete condition interface raises end-to-end success from \planned{\pDirectEnd\%} to \planned{\pFullEnd\%} for the best generator. 
Oracle studies further show that conditional video execution is the primary remaining bottleneck. 
TrafficImag provides a reproducible benchmark for evaluating and diagnosing counterfactual traffic video generation beyond perceptual video quality.
\end{abstract}

\section{Introduction}

\begin{figure*}[ht]
    \centering
    \includegraphics[width=\textwidth]{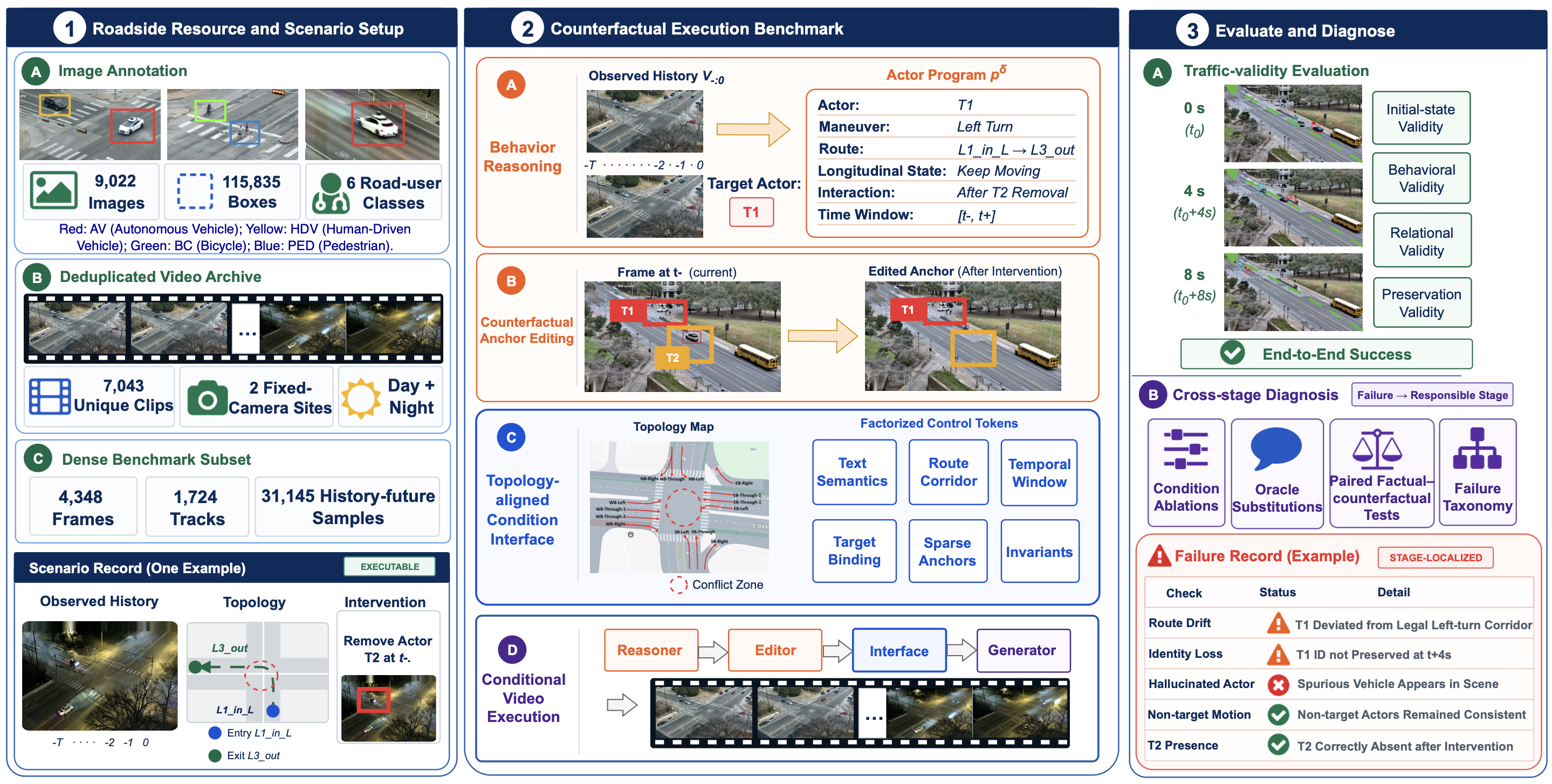}
    \caption{Overview of the TrafficImag benchmark.}
    \label{fig:pipeline}
\end{figure*}

Fixed roadside cameras observe an intersection from a stationary viewpoint and simultaneously cover approach lanes, conflict zones, crosswalks, and heterogeneous road users \cite{yehia2025arcas}. 
This global view is valuable for studying interactions that are only partially visible from an ego vehicle \cite{yehia2026egotraj,wang2026scenepilot}. 
But it also introduces substantial challenges: distant actors occupy few pixels, trajectories cross under occlusion, and the same image-plane displacement can represent different motions at different depths. 
Existing roadside datasets \cite{ye2022rope3d,yu2022dair,yu2023v2x} have supported detection, tracking, forecasting, and video question answering (VQA) from this viewpoint, but they do not directly evaluate whether a visual model can change the future of an observed scene according to a specified intervention.

A valid counterfactual roadside traffic video must satisfy two linked requirements. 
The first concerns the intervened initial state: a removed vehicle should be absent from the edited frame, an inserted vehicle should be consistent with lane orientation, perspective scale, and illumination, and all unedited regions should remain aligned with the observation. 
The second concerns the rollout: the controlled actor should retain its identity, follow a legal approach-to-exit connection, and interact consistently with nearby traffic, while actors outside the influence of the intervention should remain unchanged. 
The two requirement are coupled because an error in the location, scale, or heading of the counterfactual anchor often persists throughout the generated horizon, while a correct initial frame does not guarantee a correct future. 
For example, a generator may reintroduce a removed actor, direct the target toward an illegal exit, move a waiting vehicle without a corresponding interaction, or introduce road users that were not requested. 
Such cases can appear smooth even though they do not execute the requested counterfactual.

Current studies leave this distinction under-specified. 
On one hand, existing roadside datasets such as Rope3D \cite{ye2022rope3d}, DAIR-V2X \cite{yu2022dair}, and V2X-Seq \cite{yu2023v2x} provide geometry, detection, and trajectory supervision but do not involve generated counterfactual videos. 
SUTD-TrafficQA \cite{xu2021sutd} evaluates event and causal reasoning via VQA rather than via intervened videos. 
On the other hand, general video generation benchmarks such as VBench \cite{huang2024vbench}, T2V-CompBench \cite{sun2025t2vcompbench}, and WorldModelBench \cite{li2025worldmodelbench} assess perceptual quality, compositional alignment, and physical plausibility, but they neither bind a selected roadside actor to legal topology nor require the preservation of unrelated traffic. 
DrivingGen \cite{zhou2026drivinggen} provides evaluation of driving world model, but its principal setting is ego-view generation rather than fixed-camera actor intervention. 
As a result, high perceptual quality and valid counterfactual execution are not separately measurable in the roadside setting.

To address this gap, we introduce \textbf{TrafficImag}, a dataset and a factorized benchmark for counterfactual roadside traffic video generation (Figure \ref{fig:pipeline}). 
The main contributions are:
\begin{itemize}
    \item \textbf{A multi-layer roadside traffic resource.} We release 9,022 annotated high-resolution images with 115,835 object boxes across six road-user classes and 7,043 deduplicated video clips from two fixed-camera sites, together with an annotation-dense benchmark subset that contains 4,348 fully annotated frames, 63,543 track-linked box instances, 1,724 identity-consistent actor tracks, and 31,145 actor-centered history–future samples with trajectory-derived behavior labels and topology associations.
    \item \textbf{A counterfactual execution benchmark.} We define an evaluation protocol connecting behavior reasoning, image intervention, and conditional video generation, in which four traffic-validity dimensions and logical conjunction distinguish scene fluency from correct intervention, legal motion, coherent interaction, and non-target preservation.
    \item \textbf{A cross-stage empirical study.} We evaluate behavior reasoners, image editors, and video generators under shared scenarios, in which condition ablations, oracle substitutions, paired factual–counterfactual tests, and a failure taxonomy localize errors along the execution chain.
\end{itemize}

\section{Related Work}

\paragraph{Roadside datasets and traffic reasoning benchmarks.}
Infrastructure-view datasets provide broad spatial coverage of intersections but must handle severe scale variation and occlusion. 
Rope3D \cite{ye2022rope3d} studies monocular 3D detection from roadside images, and DAIR-V2X \cite{yu2022dair} and V2X-Seq \cite{yu2023v2x} extend infrastructure sensing to cooperative perception and sequential forecasting. 
SUTD-TrafficQA \cite{xu2021sutd} evaluates event and causal reasoning over in-the-wild traffic videos through VQA. 
DriveVLM \cite{tian2024drivevlm} and SpatialVLM \cite{chen2024spatialvlm} show that vision language models (VLMs) can connect visual evidence to driving decisions and spatial relations. 
These resources support perception and semantic reasoning, but they do not require a model to generate a modified future, nor do they provide a protocol for verifying that a generated future executes a specified intervention from a fixed roadside observation.

\begin{table*}[ht]
\centering
\footnotesize
\maxsizebox{\textwidth}{!}{%
\setlength{\tabcolsep}{3pt}
\renewcommand{\arraystretch}{1.12}
\begin{tabular}{@{}P{0.28\textwidth}*{6}{C{0.106\textwidth}}@{}}
\toprule
Benchmark & Fixed Roadside View & Generated Video & Actor Intervention & Legal Route/\allowbreak{}Topology & Non-Target Preservation & Paired Response \\
\midrule
SUTD-TrafficQA \cite{xu2021sutd} & $\checkmark$ & $\times$ & $\times$ & $\times$ & $\times$ & $\times$ \\
V2X-Seq \cite{yu2023v2x} & $\checkmark$ & $\times$ & $\times$ & $\checkmark$ & $\times$ & $\times$ \\
VBench \cite{huang2024vbench} & $\times$ & $\checkmark$ & $\times$ & $\times$ & $\times$ & $\times$ \\
T2V-CompBench \cite{sun2025t2vcompbench} & $\times$ & $\checkmark$ & $Partial$ & $\times$ & $\times$ & $\times$ \\
WorldModelBench \cite{li2025worldmodelbench} & $\times$ & $\checkmark$ & $Partial$ & $\times$ & $\times$ & $\times$ \\
DrivingGen \cite{zhou2026drivinggen} & $\times$ & $\checkmark$ & $Partial$ & $Partial$ & $\times$ & $\times$ \\
\textbf{TrafficImag (Ours)}  & $\checkmark$ & $\checkmark$ & $\checkmark$ & $\checkmark$ & $\checkmark$ & $\checkmark$ \\
\bottomrule
\end{tabular}}
\caption{Comparison of the TrafficImag benchmark (Ours) to other representative traffic and generative video benchmarks.}
\label{tab:benchmark-comparison}
\end{table*}

\begin{table*}[!t]
\centering
\footnotesize
\maxsizebox{\textwidth}{!}{%
\setlength{\tabcolsep}{3pt}
\renewcommand{\arraystretch}{1.12}
\begin{tabular}{@{}P{0.15\textwidth}P{0.17\textwidth}C{0.04\textwidth}P{0.13\textwidth}P{0.14\textwidth}P{0.12\textwidth}P{0.17\textwidth}@{}}
\toprule
Layer & Unit & Sites & Images/Frames & Boxes/Tracks & Clips/Samples & Primary Use \\
\midrule
Image Annotation Bank & PNG-XML Pair & 2 & 9,022 & 115,835 boxes & -- & Detection and Intervention Anchors \\
Video Archive & Deduplicated Clip & 2 & -- & -- & 7,043 Clips & Scene and Condition Coverage \\
Dense Benchmark Subset & Track-Linked Frame / Actor Window & 2 & 4,348 Frames & 63,543 Boxes; 1,724 Tracks & 31,145 Samples & Reasoning and Topology \\
Reasoning Audit & History-Future Sample & 2 & 25 Frames/\allowbreak{}Sample & -- & 1,500 Samples & Behavior-Program Evaluation \\
Image-Editing Suite & Intervention Case & 2 & 60 Anchors & -- & 2 Repeats/Model & Initial-State Evaluation \\
Video-Generation Suite & Counterfactual Scenario & 2 & 8 Seconds/Output & -- & 20 Scenarios; 3 Repeats & End-to-End Execution \\
\bottomrule
\end{tabular}}
\caption{TrafficImag data layers and benchmark units.}
\label{tab:dataset-summary}
\end{table*}

\paragraph{Controllable driving and traffic generation.}
GAIA-1 \cite{hu2023gaia}, DriveDreamer \cite{wang2024drivedreamer}, and Drive-WM \cite{wang2024drivewm} learn generative world models that predict future driving observations, primarily from the ego perspective.
TrafficGen \cite{feng2023trafficgen}, ProSim \cite{tan2024prosim}, and SceneDiffuser \cite{jiang2024scenediffuser} synthesize structured multi-agent traffic in trajectory space, where agent states are explicit and control is comparatively direct. 
TrailBlazer \cite{ma2024trailblazer} and FreeTraj \cite{qiu2024freetraj} bring trajectory-level control to pixel-space video generation. 
These methods are proposed as generators, however, they do not define an evaluation protocol that verifies actor-level intervention execution in pixel-space video anchored to an observed roadside scene.

\paragraph{Generative video model benchmarks.}
VBench \cite{huang2024vbench}, T2V-CompBench \cite{sun2025t2vcompbench}, WorldModelBench \cite{li2025worldmodelbench}, and DrivingGen \cite{zhou2026drivinggen} cover perceptual, compositional, physical, and driving-oriented quality. 
As summarized in Table \ref{tab:benchmark-comparison}, none of these benchmarks binds a selected actor in an observed roadside scene to legal route topology, requires the preservation of non-target traffic, or compares matched factual and counterfactual branches of the same scene.
\textbf{TrafficImag} complements this by fixing the camera viewpoint, grounding each intervention in an observed history, and jointly checking actor identity, legal route connectivity, interaction order, and non-target preservation in the same generated output.

\section{Dataset}

\subsection{Data Collection}

TrafficImag is organized into three complementary annotation layers that separate broad scene coverage from dense actor-level temporal supervision.
Table \ref{tab:dataset-summary} summarizes the three layers and the task-specific evaluation sets derived from them.
(1) The first layer is an image-annotation bank containing 9,022 high-resolution roadside images with 115,835 bounding-box annotations across six road-user classes. 
Each image has a resolution of $1920\times1080$ pixels and is paired with an XML annotation file. 
With an average of approximately 12.8 annotated road users per image, this layer provides broad coverage of actor appearance, scale, occlusion, illumination, and traffic density.
(2) The second layer is a deduplicated archive of 7,043 roadside video clips collected from two fixed-camera viewpoints, denoted SJB and Speedway, which contributes 4,177 clips and 2,866 clips, respectively. 
The archive includes 5,175 daytime and 1,868 nighttime clips, covering heterogeneous traffic densities, illumination conditions, actor scales, and occlusion patterns. 
Exact duplicates and repeated exports are removed before benchmark construction.
(3) The third layer is an annotation-dense temporal benchmark subset, curated from 31 representative clips at the two collection sites. 
It contains 4,348 fully annotated frames, 63,543 track-linked object boxes, and 1,724 identity-consistent actor tracks. 
Frames are annotated at 6 fps and capture short-term actor evolution, occlusion, interaction, and approach-to-exit motion under both daytime and nighttime conditions, with an average annotation density of 14.6 actor boxes/frame and 36.9 annotated instances/track.

From the identity-linked tracks of this subset, we construct 31,145 overlapping actor-centered history-future samples. 
Each sample contains 16 observed frames followed by 9 future frames, corresponding to approximately 2.7 s of history and 1.5 s of future at the 6 fps annotation rate, and is centered on a single visible actor rather than on an entire scene. 
Multiple actors therefore produce separate samples from the same scene-time interval, while temporally overlapping windows retain distinct observation and prediction contexts. 
On average, each track contributes approximately 18.1 actor-centered temporal samples.
Together, the three layers combine broad visual and environmental coverage with dense identity-level temporal supervision. 
The image and video layers capture variation in actor appearance, illumination, traffic density, and viewpoint, while the dense temporal layer supports actor association, behavior reasoning, topology-aware intervention, and short-horizon counterfactual evaluation.

\subsection{Data Annotation}

The image bank uses six labels: human-driven vehicle (HDV), autonomous vehicle (AV), pedestrian (PED), bicycle (BC), scooter (SCO), and motorcycle (MC). 
The dense subset links boxes into identity-consistent tracks and derives 9 behavior labels from the observed history and future of each sample: stationary, keep moving, straight trend, slowing or stop, left turn, right turn, left shift, right shift, and start moving. 
The taxonomy separates route-scale maneuvers from longitudinal state changes because the two groups differ in visual evidence and evaluation difficulty. 
Labels are multi-valued, for example, a moving actor may simultaneously exhibit a turn trend.
The class distribution and the positive-label distribution over the 31,145 samples are reported in Appendix~\ref{app:dataset-details}. 
Trajectory-derived behavior labels are treated as benchmark supervision rather than an unqualified physical ground truth. 
Labels are generated with class-aware displacement and heading rules, checked against the road topology, and manually reviewed in ambiguous cases. 
Automated quality control rejects invalid boxes, missing frame links, duplicate track identifiers, and displacements beyond class-specific thresholds. 
Manual review then verifies identity through occlusion, lane association, turn direction, and state-transition boundaries. 
This combination retains the scale of automatic processing while making the audit subset suitable for class-conditional evaluation.




\section{Benchmark}

\subsection{Task Definition}

Let $V_{-T:0}=\{I_{-T},\ldots,I_0\}$ denote the observed roadside history, $G$ the image-plane road topology, and $\delta$ an actor intervention. 
Given these inputs, the task is to produce an $H$-step future video $\hat V_{1:H}^{\delta}$ beginning from the intervened scene and executing $\delta$.
The evaluated chain is:
\begin{equation}
\begin{aligned}
P^{\delta} &= \mathcal{H}_{\omega}(V_{-T:0},G,\delta), \\
I_0^{\delta} &= \mathcal{E}_{\psi}(I_0,\delta), \\
C^{\delta} &= \mathcal{C}(G,P^{\delta},I_0^{\delta}), \\
\hat V_{1:H}^{\delta} &= \mathcal{G}_{\phi}(I_0^{\delta},C^{\delta};H),
\end{aligned}
\label{eq:pipeline}
\end{equation}
where $\mathcal{H}_{\omega}$ is a replaceable behavior reasoner, $\mathcal{E}_{\psi}$ is an image editor, $\mathcal{C}$ is the fixed benchmark condition interface, and $\mathcal{G}_{\phi}$ represents a replaceable image-to-video generator. 
This factorization enables stage-level evaluation without requiring all models to expose the same internal representation.

We evaluate two intervention types: removal and insertion. 
A removal deletes a selected actor and tests whether traffic affected by its absence responds accordingly. 
An insertion adds an actor with a specified approach, lane, and conflict relation. 
For analysis, an interacting actor is defined as one whose projected route overlaps the route of the intervened actor in a shared conflict zone within the evaluation horizon. 

\subsection{Actor Programs and Topology-Aligned Conditions}

The topology graph $G=(\mathcal L,\mathcal E_{\mathrm{con}},\mathcal Z,\Psi,\mathcal A)$ contains lane and drivable regions $\mathcal L$, legal approach-to-exit connections $\mathcal E_{\mathrm{con}}$, conflict zones $\mathcal Z$, static elements $\Psi$ such as stop lines and crosswalks, and visible actors $\mathcal A$. 
Because metric calibration is not available at both sites, lanes and routes are represented as image-plane polygons and corridors. 
These representations are adequate for testing legal connectivity and normalized motion, but are not interpreted as engineering-grade high-definition maps.
For each affected actor $i$, the reasoner returns an actor-level program record:
\begin{equation}
p_i^{\delta}=(o_i,m_i,r_i,\ell_i,\pi_i,[t_i^{-},t_i^{+}],u_i),
\end{equation}
where $o_i$ identifies the actor, $m_i$ is its maneuver, $r_i$ denotes its route or goal exit, $\ell_i$ is its longitudinal mode, $\pi_i$ is its interaction order, $[t_i^{-},t_i^{+}]$ is a temporal window, and $u_i$ represents an uncertainty field. 
The program $P^{\delta}$ collects one such record for every affected actor.
The benchmark validates this program against the topology graph before generation. 
Illegal lane-exit pairs are rejected, a removed actor cannot remain in the program, and an inserted actor must have a valid initial lane and heading. 
Actors outside the reachable intervention neighborhood of the intervention form a preserve set.
Detailed benchmark condition construction is reported in Appendix~\ref{app:benchmark}.



\subsection{Validity Dimensions}

A generated video is successful only when all four binary predicates hold:
\begin{equation}
S_{\mathrm{E2E}}=S_{\mathrm{init}}\wedge S_{\mathrm{beh}}\wedge S_{\mathrm{rel}}\wedge S_{\mathrm{pres}}.
\label{eq:validity}
\end{equation}
Initial-state validity $S_{\mathrm{init}}$ tests intervention completion, perspective, heading, lane contact, and persistence, behavioral validity $S_{\mathrm{beh}}$ tests route-corridor occupancy, arrival at the goal exit, and the programmed longitudinal state, relational validity $S_{\mathrm{rel}}$ tests yield and passage order in a shared conflict zone and rejects severe interpenetration, and preservation validity $S_{\mathrm{pres}}$ tests the state and identity of non-target actors, background stability, and unrequested actor creation or disappearance. 
The conjunction prevents a high score on one dimension from masking a critical failure on another.
The four predicates are assigned by model-blind human raters under a fixed binary rubric, with automated tracking and topology diagnostics retained alongside each decision, which are described in Appendix~\ref{app:evaluation}.

\begin{table*}[!t]
\centering
\small
\maxsizebox{\textwidth}{!}{%
\begin{tabular}{lcccccccccc}
\toprule
Model & Set F1 $\uparrow$ & Macro F1 $\uparrow$ & Rem. $\uparrow$ & Ins. $\uparrow$ & Geom. $\uparrow$ & E2E $\uparrow$ & ID $\uparrow$ & Route $\uparrow$ & Inter. $\uparrow$ & Pres. $\uparrow$ \\
\midrule
\multicolumn{11}{l}{\emph{Behavior Reasoners}} \\
\midrule
LLaVA-OneVision-7B & \planned{\pLlavaSample} & \planned{\pLlavaMacro} & -- & -- & -- & -- & -- & -- & -- & -- \\
InternVL3-14B & \planned{\pInternSample} & \planned{\pInternMacro} & -- & -- & -- & -- & -- & -- & -- & -- \\
Qwen2.5-VL-32B & \planned{\pQwenSample} & \planned{\pQwenMacro} & -- & -- & -- & -- & -- & -- & -- & -- \\
Gemini 2.5 Flash & \planned{\pGeminiFlashSample} & \planned{\pGeminiFlashMacro} & -- & -- & -- & -- & -- & -- & -- & -- \\
Claude Sonnet 4 & \planned{\pClaudeSample} & \planned{\pClaudeMacro} & -- & -- & -- & -- & -- & -- & -- & -- \\
GPT-4o & \planned{\pGPTFourOSample} & \planned{\pGPTFourOMacro} & -- & -- & -- & -- & -- & -- & -- & -- \\
Gemini 2.5 Pro & \textbf{\planned{\pGeminiProSample}} & \textbf{\planned{\pGeminiProMacro}} & -- & -- & -- & -- & -- & -- & -- & -- \\
\midrule
\multicolumn{11}{l}{\emph{Image Editors}} \\
\midrule
GPT Image 2 & -- & -- & \textbf{\planned{\pImageTwoRemoval}} & \textbf{\planned{\pImageTwoInsertion}} & \textbf{\planned{\pImageTwoGeometry}} & -- & -- & -- & -- & -- \\
Gemini 3 Pro Image & -- & -- & \planned{\pGeminiImageRemoval} & \planned{\pGeminiImageInsertion} & \planned{\pGeminiImageGeometry} & -- & -- & -- & -- & -- \\
Qwen-Image-2.0 Pro & -- & -- & \planned{\pQwenImageRemoval} & \planned{\pQwenImageInsertion} & \planned{\pQwenImageGeometry} & -- & -- & -- & -- & -- \\
Seedream 5.0 Pro & -- & -- & \planned{\pSeedreamRemoval} & \planned{\pSeedreamInsertion} & \planned{\pSeedreamGeometry} & -- & -- & -- & -- & -- \\
\midrule
\multicolumn{11}{l}{\emph{Video Generators}} \\
\midrule
Seedance 2.0 & -- & -- & -- & -- & -- & \textbf{\planned{\pSeedanceEnd}} & \textbf{\planned{\pSeedanceIdentity}} & \textbf{\planned{\pSeedanceRoute}} & \textbf{\planned{\pSeedanceInteraction}} & \textbf{\planned{\pSeedancePreservation}} \\
Sora 2 Pro & -- & -- & -- & -- & -- & \planned{\pSoraEnd} & \planned{\pSoraIdentity} & \planned{\pSoraRoute} & \planned{\pSoraInteraction} & \planned{\pSoraPreservation} \\
Veo 3.1 & -- & -- & -- & -- & -- & \planned{\pVeoEnd} & \planned{\pVeoIdentity} & \planned{\pVeoRoute} & \planned{\pVeoInteraction} & \planned{\pVeoPreservation} \\
\bottomrule
\end{tabular}}
\caption{Unified cross-model benchmark results. Behavior reasoners use 1,500 samples, image editors use 60 cases with two repeats, and video generators use 20 scenarios with three repeats. Dashes denote inapplicable metrics. Set/Macro are behavior-label F1; Rem./Ins./Geom. are edit success; E2E is the four-predicate conjunction; ID, Route, Inter., and Pres. are video validity rates. All applicable entries are percentages.}
\label{tab:unified-results}
\end{table*}

\section{Experiments}

\subsection{Experimental Setup}

The experiments address 3 main research questions (RQs):
\begin{itemize}
    \item \textbf{RQ1:} How reliably current VLMs infer the actor-level behavior program, including infrequent state transitions?
    \item \textbf{RQ2:} How do behavior reasoners, image editors, and video generators perform across the complete counterfactual execution chain?
    \item \textbf{RQ3:} Which stages and failure types limit end-to-end success, and how much can be recovered with oracle inputs or a single targeted retry?
\end{itemize}

All models are evaluated without task-specific fine-tuning. 
The open behavior reasoners are LLaVA-OneVision-7B \cite{li2024llavaonevision}, InternVL3-14B \cite{zhu2025internvl3}, and Qwen2.5-VL-32B \cite{bai2025qwen25vl}; hosted reasoners use the Gemini 2.5 \cite{google2025gemini25}, Claude Sonnet 4 \cite{anthropic2025claude4}, and GPT-4o \cite{openai2024gpt4o} families. 
The hosted image editors are GPT Image 2 \cite{openai2026gptimage2}, Gemini 3 Pro Image \cite{google2025gemini3image}, Qwen-Image-2.0 Pro \cite{zhao2026qwenimage2}, and Seedream 5.0 Pro \cite{bytedance2026seedream5pro}. 
Hosted generators were accessed on 6 July 2026. 
The recorded video identifiers are doubao-seedance-2-0-260128 \cite{bytedance2026seedance2}, sora-2-pro \cite{openai2025sora2}, and veo-3.1-generate-001 \cite{google2026veo31}. 
All outputs are eight seconds; Sora and Veo were requested at 1080p, while Seedance was requested at 720p. 
Appendix~\ref{app:config} gives all model identifiers, adapter settings, and the failure-handling policy.

Stage-specific metrics supplement the conjunction. 
Behavior reasoning uses parse success, sample-set F1, exact-set match, macro label F1, and per-label precision and recall. 
Image editing uses removal completion, insertion completion, lane-compatible geometry, non-target actor preservation, and LPIPS outside the edit mask. 
Video generation uses identity persistence, route success, interaction validity, preservation, hallucination rate, and the end-to-end conjunction. 
Perceptual measures inspired by VBench \cite{huang2024vbench} are retained as secondary diagnostics, but they do not override a failed traffic predicate. 
Paired factual-counterfactual evaluation additionally measures whether an affected actor changes entry time, waiting time, or passage order in the direction specified by the intervention.
The main results are demonstrated, and the other metrics are reported in Appendix~\ref{app:reasoning}.



\subsection{Experimental Results}

\subsubsection{RQ1: Behavior Reasoning}

Table \ref{tab:unified-results} reports the cross-model results. 
Gemini 2.5 Pro obtains the highest sample-set F1 (\planned{\pGeminiProSample\%}) and macro label F1 (\planned{\pGeminiProMacro\%}). 
The hosted models occupy the upper part of the table, while the three evaluated open VLMs range from \planned{\pLlavaMacro\%} to \planned{\pQwenMacro\%} macro F1. 
The \planned{\pReasonerSpan}-point difference between the lowest and highest macro F1 is not uniform across behaviors. 
Legal approach-to-exit connectivity and actor history provide strong evidence for turn direction, while short displacement changes, partial occlusion, and the boundary between stationary, slowing, and starting produce more disagreement across models.
The per-class audit in Appendix~\ref{app:reasoning} shows this pattern directly. 
Left- and right-turn F1 reach \planned{\pLeftTurnFOne\%} and \planned{\pRightTurnFOne\%}, respectively. 
Stationary, keep-moving, and straight-trend labels range from \planned{\pStraightFOne\%} to \planned{\pStationaryFOne\%}. 
Slowing-or-stop reaches \planned{\pSlowFOne\%}, start-moving \planned{\pStartFOne\%}, and lateral shifts \planned{\pRightShiftFOne}--\planned{\pLeftShiftFOne\%}. 
Exact-set match (\planned{\pReasonExact\%}) is lower than set F1 because a multi-label prediction can identify the dominant maneuver but miss one concurrent state label. 
These results support reporting macro and per-class measures together; parse rate or sample overlap alone would hide the remaining state-transition errors.

\subsubsection{RQ2: Cross-Stage Counterfactual Execution}

\begin{figure*}[ht]
    \centering
    \includegraphics[width=\textwidth]{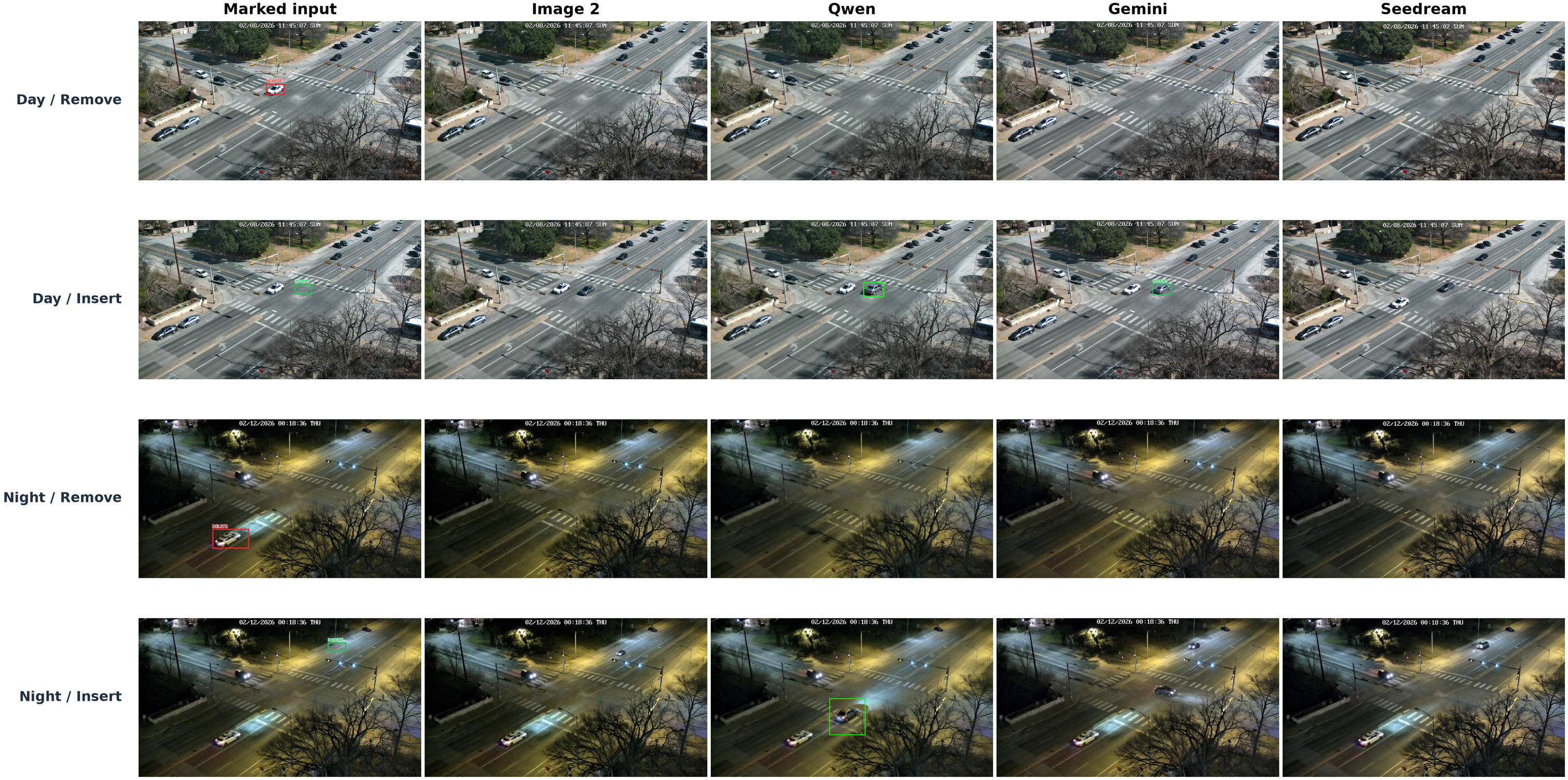}
    \caption{Day and night counterfactual anchors for vehicle removal and insertion. Each row contains the marked input followed by four editor outputs. Removal is often completed without disturbing distant traffic. Insertion exposes larger differences in target location, scale, lane contact, and illumination, especially at night.}
    \label{fig:editing}
\end{figure*}

\begin{figure*}[!t]
    \centering
    \includegraphics[width=\textwidth]{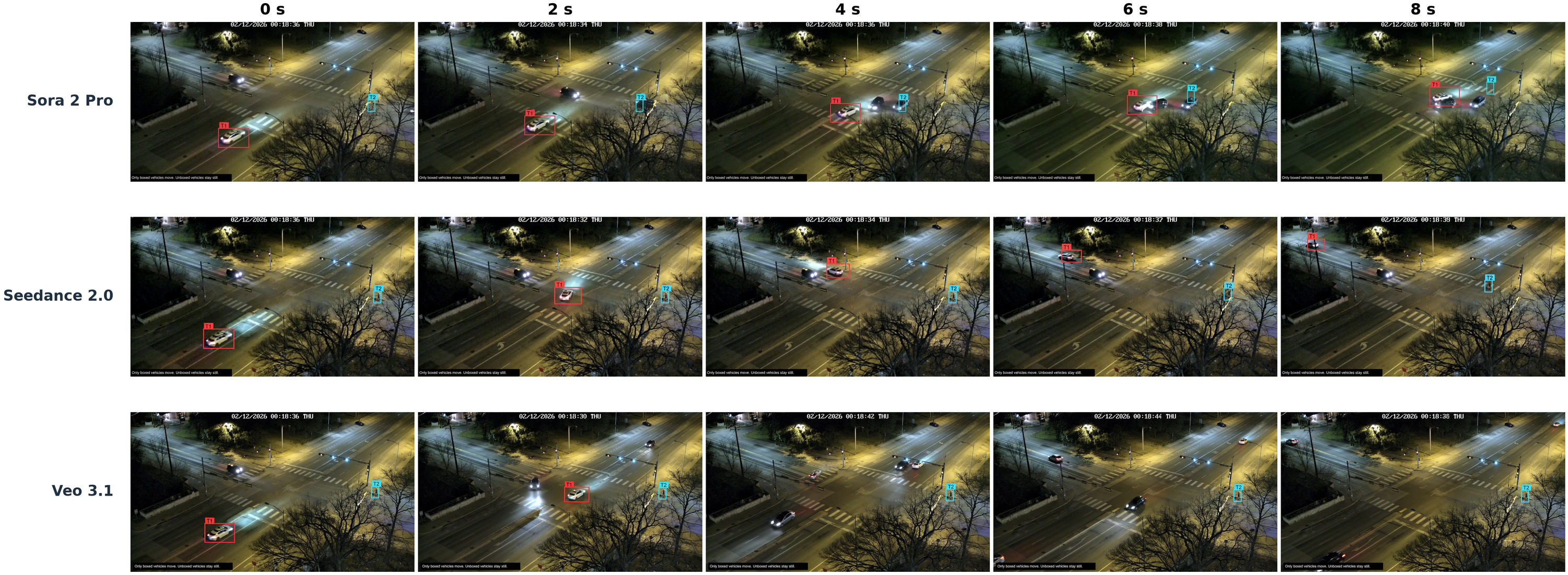}
    \caption{Matched 8-second rollouts for a programmed left-turn counterfactual. T1 is the controlled vehicle and T2 should remain stationary. Seedance completes the turn while maintaining T2 most consistently in this example. Sora retains the scene but progresses more slowly and does not complete the requested route within the shown horizon. Veo introduces additional traffic and exhibits target/path drift.}
    \label{fig:video}
\end{figure*}

The image-editing block of Table \ref{tab:unified-results} reveals a consistent difference between removal and insertion (Figure \ref{fig:editing}). 
GPT Image 2 records \planned{\pImageTwoRemoval\%} removal success and \planned{\pImageTwoInsertion\%} insertion success, and the corresponding gaps for the remaining editors range from \planned{\pEditGapMin} to \planned{\pEditGapMax} points. 
Removal mainly requires localized completion and preservation, while insertion additionally couples lane location, heading, scale, illumination, ground contact, and occlusion, so a visually integrated vehicle can still fail geometry. 
Outside-mask LPIPS remains between \planned{\pImageTwoLPIPS} and \planned{\pSeedreamLPIPS}, indicating that most failures are local intervention or geometry errors rather than global corruption.
The generator block compares a shared set of 20 scenarios under the Full interface. 
Figure \ref{fig:video} shows matched rollouts for a representative left-turn counterfactual.
Seedance 2.0 provides the strongest execution rate at \planned{\pSeedanceEnd\%}, followed by Sora 2 Pro at \planned{\pSoraEnd\%} and Veo 3.1 at \planned{\pVeoEnd\%}. 
The same ordering appears for identity, route, interaction, and preservation. 
Seedance retains the target identity in \planned{\pSeedanceIdentity\%} of outputs and reaches \planned{\pSeedanceRoute\%} route success, although some otherwise correct videos still drift in identity or route near the end. 
Sora more often follows a conservative or incomplete path, while Veo more often changes unprogrammed actors or loses the requested route. 
These stage-level scores explain why the end-to-end differences are larger than a perceptual comparison alone would suggest.
Paired factual-counterfactual results are provided in Appendix~\ref{app:component-results}.

\begin{table*}[ht]
\centering
\small
\maxsizebox{\textwidth}{!}{%
\begin{tabular}{lrrrrrr}
\toprule
Variant & Initial $\uparrow$ & Route $\uparrow$ & Interaction $\uparrow$ & Preservation $\uparrow$ & E2E $\uparrow$ & Hallucination $\downarrow$ \\
\midrule
Direct Prompt & \planned{\pDirectInit} & \planned{\pDirectRoute} & \planned{\pDirectInteraction} & \planned{\pDirectPreservation} & \planned{\pDirectEnd} & \planned{\pDirectHallucination} \\
Counterfactual Anchor & \planned{\pAnchorInit} & \planned{\pAnchorRoute} & \planned{\pAnchorInteraction} & \planned{\pAnchorPreservation} & \planned{\pAnchorEnd} & \planned{\pAnchorHallucination} \\
Anchor + Actor Program & \planned{\pTextInit} & \planned{\pTextRoute} & \planned{\pTextInteraction} & \planned{\pTextPreservation} & \planned{\pTextEnd} & \planned{\pTextHallucination} \\
Anchor + Target Binding & \planned{\pBindingInit} & \planned{\pBindingRoute} & \planned{\pBindingInteraction} & \planned{\pBindingPreservation} & \planned{\pBindingEnd} & \planned{\pBindingHallucination} \\
Full Interface & \planned{\pFullInit} & \planned{\pFullRoute} & \planned{\pFullInteraction} & \planned{\pFullPreservation} & \planned{\pFullEnd} & \planned{\pFullHallucination} \\
Full + One Repair & \textbf{\planned{\pRepairInit}} & \textbf{\planned{\pRepairRoute}} & \textbf{\planned{\pRepairInteraction}} & \textbf{\planned{\pRepairPreservation}} & \textbf{\planned{\pRepairEnd}} & \textbf{\planned{\pRepairHallucination}} \\
\bottomrule
\end{tabular}}
\caption{Incremental condition ablation with a fixed generator over 20 shared scenarios and three repeats, giving 60 outputs per variant. E2E is the conjunction of initial, behavioral, relational, and preservation validity. All entries are percentages.}
\label{tab:ablation}
\end{table*}


\subsubsection{RQ3: Bottlenecks, Failure Modes, and Repair}

Table \ref{tab:ablation} adds condition channels in execution order while holding the scenarios and the Seedance 2.0 generator fixed. 
Direct prompting reaches \planned{\pDirectEnd\%} end-to-end success. 
The counterfactual anchor raises initial-state validity from \planned{\pDirectInit\%} to \planned{\pAnchorInit\%}, but route and interaction scores change little because the future remains under-specified. 
Adding the actor program raises route success to \planned{\pTextRoute\%} and interaction validity to \planned{\pTextInteraction\%}. 
Target binding raises end-to-end success to \planned{\pBindingEnd\%} and reduces hallucination to \planned{\pBindingHallucination\%}. 
Legal corridors and scene invariants then produce the Full result of \planned{\pFullEnd\%}, a \planned{\pDirectFullGain}-point increase over direct prompting. 
Preservation rises to \planned{\pFullPreservation\%}, and hallucination falls from \planned{\pDirectHallucination\%} to \planned{\pFullHallucination\%}.
Oracle substitutions separate errors that the ablation cannot identify. 
Replacing only the predicted program with an audited program raises end-to-end success from \planned{\pFullEnd\%} to \planned{\pOracleProgramEnd\%}. 
Replacing only the edited anchor raises it to \planned{\pOracleAnchorEnd\%}, and replacing both raises it to \planned{\pOracleBothEnd\%}. 
The remaining \planned{\pGenerationGap}-point gap persists after reasoning and intervention initialization have been controlled. 
Within this pipeline, the residual is consistent with unresolved condition following, identity maintenance, and long-horizon video execution rather than an upstream program or anchor error. 
The larger anchor substitution gain also agrees with the insertion results: an initial error in position, scale, or heading is difficult for the video generator to correct later.
One-step repair is applied only to the 27 first-pass Full-interface failures. 
Four are recovered, raising end-to-end success from \planned{\pFullEnd\%} (33/60) to \planned{\pRepairEnd\%} (37/60). 
Route success rises from \planned{\pFullRoute\%} to \planned{\pRepairRoute\%}, interaction validity from \planned{\pFullInteraction\%} to \planned{\pRepairInteraction\%}, and hallucination falls from \planned{\pFullHallucination\%} to \planned{\pRepairHallucination\%}. 
The recovered cases are localized failures addressed by the selected constraint. 
The remaining cases commonly combine identity, route, and preservation errors, which explains why a single retry yields a smaller increase than the initial factorization.
Detailed failure taxonomy and the one-step repair are reported in Appendix~\ref{app:failure}.

\subsection{Discussion}

The three RQs reveal different sources of error that a single video-quality score would merge. 
\textbf{RQ1} separates structured output compliance from behavior recognition. 
Parsing is nearly complete, but state-transition labels remain less reliable than stationary motion and turn direction. 
\textbf{RQ2} then shows that correct scene initialization and correct temporal execution are distinct capabilities. 
Removal is mainly a localized completion problem, whereas insertion additionally requires perspective, lane contact, scale, and occlusion to agree. 
Even a correct anchor does not prevent later identity or route drift. 
\textbf{RQ3} further shows that these errors propagate across stages: improving the program or the anchor raises downstream success, but neither removes failures that emerge during the rollout.

The oracle results make this propagation explicit. 
Auditing the anchor produces a larger gain than auditing the program alone, which is consistent with the image-editing results and with the difficulty of recovering from an incorrect initial pose. 
Auditing both inputs raises success further yet does not close the remaining gap. 
The benchmark should therefore not be interpreted as a leaderboard for isolated generators. 
It measures an execution chain in which program inference, intervention initialization, condition translation, and temporal generation can each determine the final outcome. 
Reporting the predicate vector and dominant failure label alongside end-to-end success preserves this distinction.

Paired factual–counterfactual evaluation addresses a limitation of output-only assessment.
Two branches can each be visually plausible while showing no response to the intervention, so intervention-response consistency tests a relation between matched outputs rather than the quality of either video in isolation. 
Additionally, hosted generators differ in reference-image limits, resolution, seed exposure, and prompt handling.
The benchmark fixes the semantic program, records each adapter transformation, and retains failed calls, so the reported results describe performance under a disclosed protocol and access date rather than a timeless model ranking. 
The failure records also indicate concrete model-development targets: persistent actor binding, explicit route memory, preserve-set constraints, and uncertainty-aware handling of ambiguous state transitions.

\section{Conclusion}

TrafficImag introduces an execution-oriented benchmark for counterfactual roadside traffic video generation, decomposing the task into behavior reasoning, intervention initialization, conditional video generation, and traffic-aware verification, so that failures are localized rather than hidden behind aggregate video-quality metrics. 
It grounds this protocol in a large roadside image and video corpus with a densely tracked benchmark subset.
Four explicit validity dimensions expose errors that perceptual fluency alone cannot identify. 
Across the evaluated models, topology-aligned conditions improve execution, while oracle and paired tests isolate persistent limitations in anchor geometry, long-horizon condition following, and coupled multi-actor behavior. 
By releasing scenario definitions, adapters, predicate-level scores, and failure records, the benchmark provides a reproducible basis for measuring progress beyond visually plausible continuation.

Several limitations qualify these results.
TrafficImag covers two fixed-camera sites and does not represent the full diversity of road rules, camera elevations, weather, and intersection geometry, and because topology is expressed in normalized image coordinates, the protocol tests legal connectivity, relative displacement, and interaction order but cannot certify metric speed, clearance, or collision risk. 
Behavior labels derived from short trajectories remain ambiguous near the boundary between stationary, slowing, and starting, and the 20-scenario generation suite supports paired diagnosis but requires larger scenario sets for narrower confidence intervals and for conclusions about geographic or geometric generalization. 
The benchmark evaluates conditional association with a specified intervention rather than causal identification from observational data, and hosted-model behavior may change after provider updates, so exact identifiers, access dates, and payloads form part of the result record. 
TrafficImag is intended for model evaluation, visualization, and failure analysis, not for signal control, safety certification, or the replacement of calibrated traffic simulation.
Extending the protocol to additional calibrated sites and evaluating generators designed for persistent actor binding and explicit route memory are natural directions for future work.

{\footnotesize
\bibliography{references}

@inproceedings{ye2022rope3d,
  title={Rope3d: The roadside perception dataset for autonomous driving and monocular 3d object detection task},
  author={Ye, Xiaoqing and Shu, Mao and Li, Hanyu and Shi, Yifeng and Li, Yingying and Wang, Guangjie and Tan, Xiao and Ding, Errui},
  booktitle={Proceedings of the IEEE/CVF Conference on Computer Vision and Pattern Recognition},
  pages={21341--21350},
  year={2022}
}

@inproceedings{yu2022dair,
  title={Dair-v2x: A large-scale dataset for vehicle-infrastructure cooperative 3d object detection},
  author={Yu, Haibao and Luo, Yizhen and Shu, Mao and Huo, Yiyi and Yang, Zebang and Shi, Yifeng and Guo, Zhenglong and Li, Hanyu and Hu, Xing and Yuan, Jirui and others},
  booktitle={Proceedings of the IEEE/CVF conference on computer vision and pattern recognition},
  pages={21361--21370},
  year={2022}
}

@inproceedings{yu2023v2x,
  title={V2x-seq: A large-scale sequential dataset for vehicle-infrastructure cooperative perception and forecasting},
  author={Yu, Haibao and Yang, Wenxian and Ruan, Hongzhi and Yang, Zhenwei and Tang, Yingjuan and Gao, Xu and Hao, Xin and Shi, Yifeng and Pan, Yifeng and Sun, Ning and others},
  booktitle={Proceedings of the IEEE/CVF Conference on Computer Vision and Pattern Recognition},
  pages={5486--5495},
  year={2023}
}

@inproceedings{xu2021sutd,
  title={Sutd-trafficqa: A question answering benchmark and an efficient network for video reasoning over traffic events},
  author={Xu, Li and Huang, He and Liu, Jun},
  booktitle={Proceedings of the IEEE/CVF conference on computer vision and pattern recognition},
  pages={9878--9888},
  year={2021}
}

@article{tian2024drivevlm,
  title={Drivevlm: The convergence of autonomous driving and large vision-language models},
  author={Tian, Xiaoyu and Gu, Junru and Li, Bailin and Liu, Yicheng and Wang, Yang and Zhao, Zhiyong and Zhan, Kun and Jia, Peng and Lang, Xianpeng and Zhao, Hang},
  journal={arXiv preprint arXiv:2402.12289},
  year={2024}
}

@inproceedings{chen2024spatialvlm,
  title={Spatialvlm: Endowing vision-language models with spatial reasoning capabilities},
  author={Chen, Boyuan and Xu, Zhuo and Kirmani, Sean and Ichter, Brain and Sadigh, Dorsa and Guibas, Leonidas and Xia, Fei},
  booktitle={Proceedings of the IEEE/CVF Conference on Computer Vision and Pattern Recognition},
  pages={14455--14465},
  year={2024}
}

@article{hu2023gaia,
  title={Gaia-1: A generative world model for autonomous driving},
  author={Hu, Anthony and Russell, Lloyd and Yeo, Hudson and Murez, Zak and Fedoseev, George and Kendall, Alex and Shotton, Jamie and Corrado, Gianluca},
  journal={arXiv preprint arXiv:2309.17080},
  year={2023}
}

@inproceedings{wang2024drivedreamer,
  title={Drivedreamer: Towards real-world-drive world models for autonomous driving},
  author={Wang, Xiaofeng and Zhu, Zheng and Huang, Guan and Chen, Xinze and Zhu, Jiagang and Lu, Jiwen},
  booktitle={European conference on computer vision},
  pages={55--72},
  year={2024},
  organization={Springer}
}

@inproceedings{wang2024drivewm,
  title={Driving into the future: Multiview visual forecasting and planning with world model for autonomous driving},
  author={Wang, Yuqi and He, Jiawei and Fan, Lue and Li, Hongxin and Chen, Yuntao and Zhang, Zhaoxiang},
  booktitle={Proceedings of the IEEE/CVF Conference on Computer Vision and Pattern Recognition},
  pages={14749--14759},
  year={2024}
}

@article{feng2023trafficgen,
  title={Trafficgen: Learning to generate diverse and realistic traffic scenarios},
  author={Feng, Lan and Li, Quanyi and Peng, Zhenghao and Tan, Shuhan and Zhou, Bolei},
  journal={arXiv preprint arXiv:2210.06609},
  year={2022}
}

@article{tan2024prosim,
  title={Promptable closed-loop traffic simulation},
  author={Tan, Shuhan and Ivanovic, Boris and Chen, Yuxiao and Li, Boyi and Weng, Xinshuo and Cao, Yulong and Kr{\"a}henb{\"u}hl, Philipp and Pavone, Marco},
  journal={arXiv preprint arXiv:2409.05863},
  year={2024}
}

@inproceedings{jiang2024scenediffuser,
  title={Scenediffuser: Efficient and controllable driving simulation initialization and rollout},
  author={Jiang, Chiyu Max and Bai, Yijing and Cornman, Andre and Davis, Christopher and Huang, Xiukun and Jeon, Hong and Kulshrestha, Sakshum and Lambert, John Wheatley and Li, Shuangyu and Zhou, Xuanyu and others},
  booktitle={The Thirty-eighth Annual Conference on Neural Information Processing Systems},
  year={2024}
}

@article{ma2024trailblazer,
  title={Trailblazer: Trajectory control for diffusion-based video generation},
  author={Ma, Wan-Duo Kurt and Lewis, John P and Kleijn, W Bastiaan},
  journal={arXiv preprint arXiv:2401.00896},
  year={2023}
}

@article{qiu2024freetraj,
  title={Freetraj: Tuning-free trajectory control in video diffusion models},
  author={Qiu, Haonan and Chen, Zhaoxi and Wang, Zhouxia and He, Yingqing and Xia, Menghan and Liu, Ziwei},
  journal={arXiv preprint arXiv:2406.16863},
  year={2024}
}

@inproceedings{huang2024vbench,
  title={Vbench: Comprehensive benchmark suite for video generative models},
  author={Huang, Ziqi and He, Yinan and Yu, Jiashuo and Zhang, Fan and Si, Chenyang and Jiang, Yuming and Zhang, Yuanhan and Wu, Tianxing and Jin, Qingyang and Chanpaisit, Nattapol and others},
  booktitle={Proceedings of the IEEE/CVF Conference on Computer Vision and Pattern Recognition},
  pages={21807--21818},
  year={2024}
}

@inproceedings{sun2025t2vcompbench,
  title={T2v-compbench: A comprehensive benchmark for compositional text-to-video generation},
  author={Sun, Kaiyue and Huang, Kaiyi and Liu, Xian and Wu, Yue and Xu, Zihan and Li, Zhenguo and Liu, Xihui},
  booktitle={Proceedings of the Computer Vision and Pattern Recognition Conference},
  pages={8406--8416},
  year={2025}
}

@article{li2025worldmodelbench,
  title={Worldmodelbench: Judging video generation models as world models},
  author={Li, Dacheng and Fang, Yunhao and Chen, Yukang and Yang, Shuo and Cao, Shiyi and Wong, Justin and Luo, Michael and Wang, Xiaolong and Yin, Hongxu and Gonzalez, Joseph and others},
  journal={Advances in Neural Information Processing Systems},
  volume={38},
  year={2026}
}

@article{zhou2026drivinggen,
  title={DrivingGen: A Comprehensive Benchmark for Generative Video World Models in Autonomous Driving},
  author={Zhou, Yang and Shao, Hao and Wang, Letian and Zong, Zhuofan and Li, Hongsheng and Waslander, Steven L},
  journal={arXiv preprint arXiv:2601.01528},
  year={2026}
}

@article{li2024llavaonevision,
  title={Llava-onevision: Easy visual task transfer},
  author={Li, Bo and Zhang, Yuanhan and Guo, Dong and Zhang, Renrui and Li, Feng and Zhang, Hao and Zhang, Kaichen and Zhang, Peiyuan and Li, Yanwei and Liu, Ziwei and others},
  journal={arXiv preprint arXiv:2408.03326},
  year={2024}
}

@article{zhu2025internvl3,
  title={Internvl3: Exploring advanced training and test-time recipes for open-source multimodal models},
  author={Zhu, Jinguo and Wang, Weiyun and Chen, Zhe and Liu, Zhaoyang and Ye, Shenglong and Gu, Lixin and Tian, Hao and Duan, Yuchen and Su, Weijie and Shao, Jie and others},
  journal={arXiv preprint arXiv:2504.10479},
  year={2025}
}

@misc{bai2025qwen25vl,
      title={Qwen2.5-VL Technical Report}, 
      author={Shuai Bai and Keqin Chen and Xuejing Liu and Jialin Wang and Wenbin Ge and Sibo Song and Kai Dang and Peng Wang and Shijie Wang and Jun Tang and Humen Zhong and Yuanzhi Zhu and Mingkun Yang and Zhaohai Li and Jianqiang Wan and Pengfei Wang and Wei Ding and Zheren Fu and Yiheng Xu and Jiabo Ye and Xi Zhang and Tianbao Xie and Zesen Cheng and Hang Zhang and Zhibo Yang and Haiyang Xu and Junyang Lin},
      year={2025},
      eprint={2502.13923},
      archivePrefix={arXiv},
      primaryClass={cs.CV},
      url={https://arxiv.org/abs/2502.13923}, 
}

@article{zhao2026qwenimage2,
  title={Qwen-image-2.0 technical report},
  author={Zhao, Bing and Wu, Chenfei and Li, Deqing and Meng, Hao and Li, Jiahao and Zhang, Jie and Zhou, Jingren and Lin, Junyang and Gao, Kaiyuan and Cao, Kuan and others},
  journal={arXiv preprint arXiv:2605.10730},
  year={2026}
}

@article{openai2024gpt4o,
  title={Gpt-4o system card},
  author={Hurst, Aaron and Lerer, Adam and Goucher, Adam P and Perelman, Adam and Ramesh, Aditya and Clark, Aidan and Ostrow, AJ and Welihinda, Akila and Hayes, Alan and Radford, Alec and others},
  journal={arXiv preprint arXiv:2410.21276},
  year={2024}
}

@article{anthropic2025claude4,
  title={System card: Claude opus 4 \& claude sonnet 4},
  author={Anthropic, AI},
  journal={Claude-4 Model Card},
  year={2025}
}

@misc{google2025gemini25,
      title={Gemini 2.5: Pushing the Frontier with Advanced Reasoning, Multimodality, Long Context, and Next Generation Agentic Capabilities}, 
      author={Gheorghe Comanici and et al.},
      year={2025},
      eprint={2507.06261},
      archivePrefix={arXiv},
      primaryClass={cs.CL},
      url={https://arxiv.org/abs/2507.06261}, 
}

@misc{openai2025sora2,
  title={Sora 2 System Card},
  author={{OpenAI}},
  howpublished={Official OpenAI system card},
  year={2025},
  url={https://openai.com/index/sora-2-system-card/}
}

@misc{google2026veo31,
  title={Introducing Veo 3.1 and Advanced Capabilities in Flow},
  author={{Google}},
  howpublished={Official Google product announcement},
  year={2025},
  url={https://blog.google/innovation-and-ai/products/veo-updates-flow/}
}

@misc{bytedance2026seedance2,
  title={Seedance 2.0: Advancing video generation for world complexity},
  author={Seedance, Team and Chen, De and Chen, Liyang and Chen, Xin and Chen, Ying and Chen, Zhuo and Chen, Zhuowei and Cheng, Feng and Cheng, Tianheng and Cheng, Yufeng and others},
  journal={arXiv preprint arXiv:2604.14148},
  year={2026}
}

@misc{openai2026gptimage2,
  title={GPT Image 2 Model},
  author={{OpenAI}},
  howpublished={Official OpenAI API documentation},
  year={2026},
  url={https://developers.openai.com/api/docs/models/gpt-image-2}
}

@misc{google2025gemini3image,
  title={Gemini 3 Pro Image: Model Card},
  author={{Google DeepMind}},
  howpublished={Official Google DeepMind model card},
  year={2025},
  url={https://storage.googleapis.com/deepmind-media/Model-Cards/Gemini-3-Pro-Image-Model-Card.pdf}
}

@misc{bytedance2026seedream5pro,
  title={Seedream 5.0 Pro},
  author={{ByteDance Seed Team}},
  howpublished={Official ByteDance Seed model page},
  year={2026},
  url={https://seed.bytedance.com/en/seedream5_0_pro}
}

@article{yehia2026egotraj,
  title={EgoTraj: Real-World Egocentric Human Trajectory Dataset for Multimodal Prediction},
  author={Yehia, Ahmad and Mohamed, Abduallah and Wang, Tianyi and Byeon, Jiseop and Qian, Kun and Jiao, Junfeng and Claudel, Christian},
  journal={arXiv preprint arXiv:2605.19004},
  year={2026}
}

@article{yehia2025arcas,
  title={ARCAS: An Augmented Reality Collision Avoidance System with SLAM-Based Tracking for Enhancing VRU Safety},
  author={Yehia, Ahmad and Byeon, Jiseop and Wang, Tianyi and Wang, Huihai and Xu, Yiming and Jiao, Junfeng and Claudel, Christian},
  journal={arXiv preprint arXiv:2512.05299},
  year={2025}
}

@article{wang2026scenepilot,
  title={ScenePilot-Bench: A Large-Scale Dataset and Benchmark for Evaluation of Vision-Language Models in Autonomous Driving},
  author={Wang, Yujin and Zheng, Yutong and Fan, Wenxian and Wang, Tianyi and Chu, Hongqing and Tian, Daxin and Gao, Bingzhao and Wang, Jianqiang and Chen, Hong},
  journal={arXiv preprint arXiv:2601.19582},
  year={2026}
}
}

\newpage
\appendix
\onecolumn
\setcounter{secnumdepth}{1}

\section{Dataset Statistics and Evaluation Subsets}
\label{app:dataset-details}

\subsection{Dataset Layers and Class Composition}
TrafficImag contains an image-annotation bank, a deduplicated video archive, and a dense temporal subset. Table~\ref{tab:class-statistics} reports image-bank boxes and dense-subset tracks. The image bank is reported in terms of annotated bounding boxes, whereas the dense temporal subset is reported in terms of object tracks. Because these statistics measure different quantities, they are intended to characterize the two subsets independently and should not be compared directly.

\begin{center}
\begin{minipage}{\textwidth}
\centering
\small
\begin{tabular}{lrr}
\toprule
Class & Image Boxes & Dense Tracks \\
\midrule
HDV & 81,963 & 868 \\
AV  & 8,101  & 78 \\
PED & 19,454 & 578 \\
BC  & 3,162  & 106 \\
SCO & 2,416  & 62 \\
MC  & 739    & 32 \\
\midrule
Total & 115,835 & 1,724 \\
\bottomrule
\end{tabular}
\normalsize
\captionof{table}{Class distribution in the image-annotation bank and dense temporal subset.}
\label{tab:class-statistics}
\end{minipage}
\end{center}

Actor-centered samples use the 16 observed frames and 9 future frames defined in the main paper. Samples are constructed for individual actors rather than entire scenes; therefore, multiple actors appearing in the same video segment generate separate samples. Overlapping windows are retained when either the observation interval or the prediction interval differs.

\subsection{Behavior-Label Distribution}
Behavior labels are not mutually exclusive. Across the 31,145 samples, the positive counts are 18,871 stationary, 10,582 keep moving, 13,995 straight trend, 1,300 slowing or stop, 8,345 left turn, 8,805 right turn, 906 left shift, 753 right shift, and 417 start moving. The corresponding percentages do not sum to 100\%.

Each sample has one primary route-scale label. Longitudinal labels are mutually exclusive in most cases; 25 transition windows retain two adjacent states because the state changes within the prediction interval. Lateral shifts are treated independently and may occur simultaneously with any longitudinal behavior. We therefore report sample-set F1, exact-set match, macro label F1, and per-label precision and recall. Figure~\ref{fig:dataset-distributions} summarizes the class, behavior-label, site, and illumination distributions.

\begin{center}
\begin{minipage}{\textwidth}
    \centering
    \includegraphics[width=\textwidth]{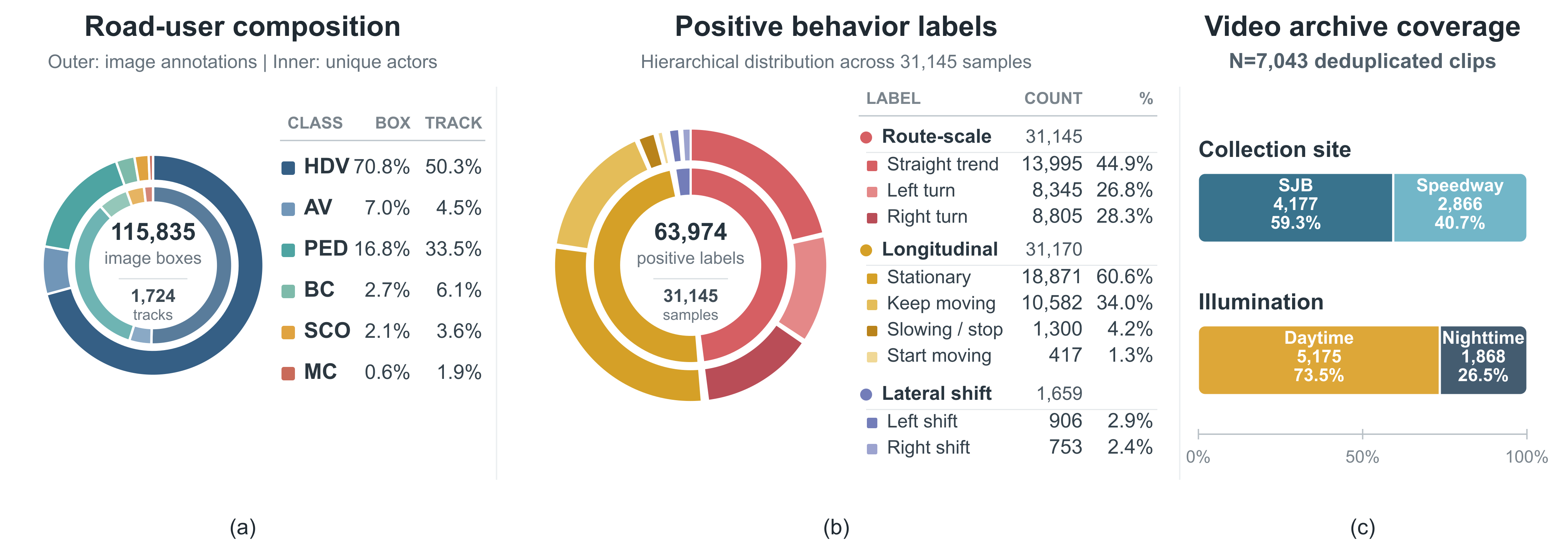}
    \captionof{figure}{TrafficImag data composition. (a) Road-user classes in the image-annotation bank (outer ring; 115,835 boxes) and dense temporal subset (inner ring; 1,724 tracks). (b) Positive behavior-label counts over 31,145 actor-centered samples; labels are multi-valued. (c) The 7,043 deduplicated clips by site and illumination.}
    \label{fig:dataset-distributions}
\end{minipage}
\end{center}

\subsection{Evaluation Subsets and Accounting}
Table~\ref{tab:evaluation-accounting} separates benchmark cases, repeats, and repair calls. The Full~+~One Repair result in Table~\ref{tab:ablation} uses the same 60 Full-interface slots; only the 27 failed slots receive a second request.

\begin{center}
\begin{minipage}{\textwidth}
\centering
\small
\maxsizebox{\textwidth}{!}{%
\begin{tabular}{llllr}
\toprule
Evaluation Block & Unit & Cases or Scenarios & Repeats & Scored Records / Additional Calls \\
\midrule
Behavior reasoning & Actor-centered sample & 1,500 per model & 1 & 1,500 per model \\
Image editing & Intervention case & 60 per editor & 2 & 120 per editor \\
Generator comparison & Counterfactual scenario & 20 per generator & 3 & 60 per generator \\
Condition ablation & Scenario-condition pair & 20 $\times$ 5 first-pass conditions & 3 & 300 first-pass outputs \\
One-step repair & Failed Full-interface slot & 27 failures & 1 & 27 additional calls; 60 final slot decisions \\
Paired response & Factual and counterfactual pair & 15 conflict scenarios & 3 & 45 pairs / 90 videos \\
\bottomrule
\end{tabular}}
\normalsize
\captionof{table}{Evaluation subsets and accounting. A slot denotes one fixed position for a scenario and repeat.}
\label{tab:evaluation-accounting}
\end{minipage}
\end{center}

Models within each benchmark block use the same cases. The reasoning audit fixes the histories, target identifiers, topology, and label schema; the editing suite contains 30 removal and 30 insertion cases; and the video suite contains 20 scenarios with three repeats. Failed calls and malformed or incorrectly sized outputs remain in the denominator. Geometry is reported in normalized image coordinates unless calibration is available.

\section{Benchmark Condition Construction}
\label{app:benchmark}

\subsection{Actor-Program Schema}
Each affected actor is represented by the program record in Equation~(2). Table~\ref{tab:program-schema} lists the fields and consistency checks. Model-specific prompts are generated from this record.

\begin{center}
\begin{minipage}{\textwidth}
\centering
\small
\maxsizebox{\textwidth}{!}{%
\begin{tabular}{llll}
\toprule
Field & Symbol & Meaning & Consistency Check \\
\midrule
Actor binding & $o_i$ & Target or interacting actor identity & Must match an observed actor or the inserted anchor \\
Maneuver & $m_i$ & Route-scale behavior, such as left turn or straight & Must be compatible with the selected route \\
Route / goal & $r_i$ & Legal approach-to-exit connection & Lane-to-exit pair must exist in the topology graph \\
Longitudinal mode & $\ell_i$ & Stationary, moving, slowing, or starting state & Must agree with the temporal instruction \\
Interaction order & $\pi_i$ & Yield or passage order in a shared conflict zone & Must not contradict another actor record \\
Temporal window & $[t_i^{-},t_i^{+}]$ & Intended execution interval & Must lie within the generated horizon \\
Uncertainty & $u_i$ & Ambiguity or tolerance retained for audit & Recorded for audit; does not override the discrete program fields \\
\bottomrule
\end{tabular}}
\normalsize
\captionof{table}{Actor-program fields used by the condition compiler.}
\label{tab:program-schema}
\end{minipage}
\end{center}

\subsection{Topology-Aware Compilation}
The compiler takes the image-plane topology, actor program, and edited anchor as input. It produces five groups of conditions:
\begin{enumerate}
    \item \textbf{Visual anchor:} the edited first frame at time zero;
    \item \textbf{Semantic program:} controlled actors, maneuvers, longitudinal states, and interaction order;
    \item \textbf{Target binding:} actor identifiers and image-plane boxes;
    \item \textbf{Route and anchors:} a legal route corridor and sparse entry, conflict-zone, curvature, and exit anchors;
    \item \textbf{Time and invariants:} temporal windows, fixed-camera constraints, preserve-set actors, persistent removal, and prohibition of unrequested actor creation.
\end{enumerate}

Before generation, the compiler rejects invalid lane-to-exit pairs, removal programs that retain the removed actor, lane-incompatible insertion headings, out-of-range temporal windows, and contradictory interaction orders. Actors outside the intervention neighborhood form the preserve set. Because the sites lack a shared metric calibration, corridors and anchors use normalized image coordinates and support legal-route and relative-motion checks only.

Figure~\ref{fig:conditions} illustrates how the common
specification is compiled for video generation.

\begin{center}
\begin{minipage}{\textwidth}
    \centering
    \includegraphics[width=\textwidth]{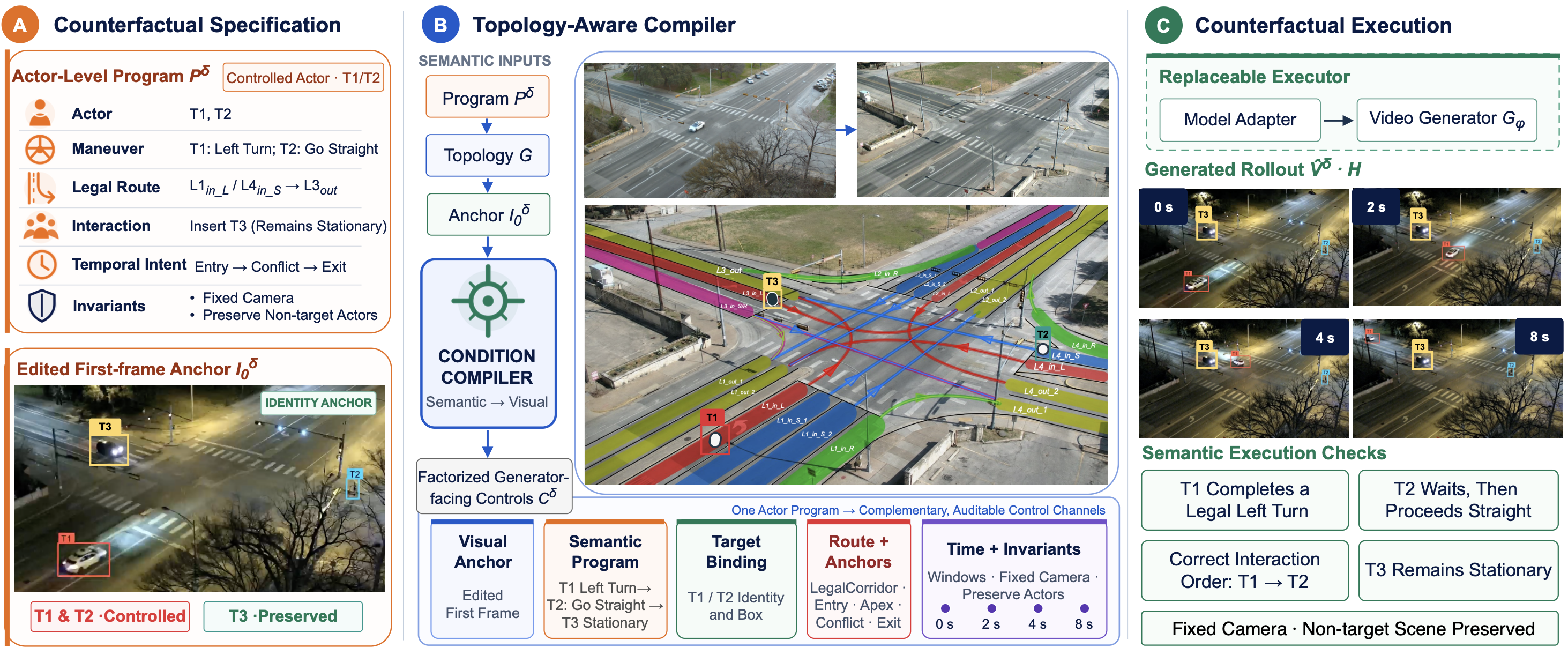}
    \captionof{figure}{Condition construction for one counterfactual scenario. The compiler maps the actor program, edited anchor, and image-plane topology to target bindings, route constraints, sparse anchors, timing, and scene invariants. Adapters translate the specification into each service's supported input format.}
    \label{fig:conditions}
\end{minipage}
\end{center}

When a service accepts multiple images, the anchor and topology
overlay are supplied separately. Otherwise, the visual cues are combined in one marked reference, and route and timing constraints are restated in text.

\section{Evaluation Rubric}
\label{app:evaluation}

\subsection{Four Binary Validity Predicates}
Three raters, blind to model identity, evaluate outputs presented in randomized order. For each output, they inspect the intervention, edited anchor, target bindings, route overlay, generated video, and preserve-set tracks, and assign the four binary predicates in Table~\ref{tab:rubric}. End-to-end success requires all four predicates.

\begin{center}
\begin{minipage}{\textwidth}
\centering
\small
\maxsizebox{\textwidth}{!}{%
\begin{tabular}{P{0.12\textwidth}P{0.32\textwidth}P{0.36\textwidth}P{0.16\textwidth}}
\toprule
Predicate & Pass Condition & Hard-Failure Examples & Retained Diagnostics \\
\midrule
$S_{\mathrm{init}}$ & The requested removal or insertion is completed; inserted geometry is consistent with lane orientation, perspective, and road contact; and the intervention persists. & Missing intervention; reappearance of a removed actor; or incompatible insertion location, scale, heading, or lane contact. & Edit completion, box geometry, target-track duration. \\
$S_{\mathrm{beh}}$ & The controlled actor follows the legal route corridor, reaches the programmed exit when required, and follows the requested longitudinal mode. & Illegal exit, route drift, incorrect maneuver, or failure to execute the programmed motion within the horizon. & Corridor occupancy, route-anchor progression, goal-exit match. \\
$S_{\mathrm{rel}}$ & Actors sharing a conflict zone satisfy the programmed yield or passage order without severe interpenetration. & Reversed order, missing response by an affected actor, or severe overlap inconsistent with the program. & Conflict-zone entry time, passage order, overlap diagnostics. \\
$S_{\mathrm{pres}}$ & Non-target actor identity and state remain consistent; the camera and background remain stable; no persistent unrequested actor appears; and no non-target actor disappears. & Unexplained preserve-set motion, identity loss, persistent actor creation, non-target disappearance, or camera/background drift. & Preserve-set displacement, unmatched tracks, background and outside-mask diagnostics. \\
\bottomrule
\end{tabular}}
\normalsize
\captionof{table}{Operational rubric for the four traffic-validity predicates.}
\label{tab:rubric}
\end{minipage}
\end{center}

Texture flicker is recorded separately. Traffic validity is determined by the four predicates, with tracking and topology projections used only as diagnostic evidence.

\subsection{Rater and Statistical Protocol}
Raters assess the time-zero intervention, target identity and route, interaction order, and preserve-set behavior in that order. Disagreements are adjudicated from stored frames, tracks, and topology without revealing model identity.

For video-generation and paired evaluations, confidence intervals use a scenario-level cluster bootstrap; repeats and paired branches
from the same scenario remain in the same cluster. The behavior-reasoning audit uses the clip-level bootstrap described in Appendix~\ref{app:reasoning}. Direct and Full conditions are paired by scenario, and failed calls remain in the denominator.

\section{Model Adapters and Execution Policy}
\label{app:config}

\subsection{Model Identifiers}

Tables~\ref{tab:reasoner-editor-config}
and~\ref{tab:generator-config} list the model identifiers and
video-generation settings used in the benchmark.

\begin{center}
\begin{minipage}{\textwidth}
\centering
\small
\maxsizebox{\textwidth}{!}{%
\begin{tabular}{lll}
\toprule
Stage & Short name & Weight repository or service identifier \\
\midrule
Reasoner & LLaVA-OneVision-7B & \texttt{lmms-lab/llava-onevision-qwen2-7b-ov} \\
Reasoner & InternVL3-14B & \texttt{OpenGVLab/InternVL3-14B} \\
Reasoner & Qwen2.5-VL-32B & \texttt{Qwen/Qwen2.5-VL-32B-Instruct} \\
Reasoner & Gemini 2.5 Flash & \texttt{gemini-2.5-flash} \\
Reasoner & Claude Sonnet 4 & \texttt{claude-sonnet-4-20250514} \\
Reasoner & GPT-4o & \texttt{gpt-4o-2024-08-06} \\
Reasoner & Gemini 2.5 Pro & \texttt{gemini-2.5-pro} \\
\midrule
Editor & GPT Image 2 & \texttt{gpt-image-2} \\
Editor & Gemini 3 Pro Image & \texttt{gemini-3-pro-image-preview} \\
Editor & Qwen-Image-2.0 Pro & \texttt{qwen-image-2.0-pro} \\
Editor & Seedream 5.0 Pro & \texttt{doubao-seedream-5-0-pro-260628} \\
\bottomrule
\end{tabular}}
\normalsize
\captionof{table}{Reasoner and image-editor identifiers used in the benchmark. Hosted access dates are retained with the execution records.}
\label{tab:reasoner-editor-config}
\end{minipage}
\end{center}

\begin{center}
\begin{minipage}{\textwidth}
\centering
\small
\maxsizebox{\textwidth}{!}{%
\begin{tabular}{llllll}
\toprule
Short name & Model identifier & Duration & Requested output & Visual format & Repeats \\
\midrule
Seedance 2.0 & \texttt{doubao-seedance-2-0-260128} & 8 s & 720p & 16:9 first-frame reference & 3 \\
Sora 2 Pro & \texttt{sora-2-pro} & 8 s & 1080p & $1280\times720$ input canvas & 3 \\
Veo 3.1 & \texttt{veo-3.1-generate-001} & 8 s & 1080p & 16:9 first-frame reference & 3 \\
\bottomrule
\end{tabular}}
\normalsize
\captionof{table}{Recorded configuration for the aggregate video-generation benchmark. Provider-side hardware is not exposed.}
\label{tab:generator-config}
\end{minipage}
\end{center}

No model is fine-tuned for the benchmark. Reasoners receive identical histories, target identifiers, topology, label vocabularies, and output schemas. Unexposed parameters are recorded as unavailable. Each video execution record stores the model and request metadata, image order, requested settings, seed when available, latency, returned dimensions, status, and error code.

\subsection{Common Semantic Serialization}
Table~\ref{tab:adapter-serialization} lists the common condition fields serialized by each adapter. Wording changes only when required by a service interface.

\begin{center}
\begin{minipage}{\textwidth}
\centering
\small
\maxsizebox{\textwidth}{!}{%
\begin{tabular}{lll}
\toprule
Condition Field & Generator-Facing Representation & Example Content \\
\midrule
Scene and horizon & Opening instruction & Fixed roadside camera; realistic 8-second continuation \\
Target binding & Actor ID plus image-plane box or marked reference & T1 is controlled; T2 is interacting; unbound actors are preserved \\
Route semantics & Legal corridor and ordered sparse anchors & Entry $\rightarrow$ conflict zone $\rightarrow$ exit \\
Temporal program & Relative start and passage windows & T2 waits; T1 enters first; T2 proceeds afterward \\
Longitudinal behavior & Actor-specific motion state & Stationary, keep moving, slow or stop, or start moving \\
Invariants & Explicit negative constraints & Fixed camera; unchanged road; preserve actors; no new actors \\
\bottomrule
\end{tabular}}
\normalsize
\captionof{table}{Common semantic fields serialized by each video-model adapter.}
\label{tab:adapter-serialization}
\end{minipage}
\end{center}

The reference image defines time zero. Textual coordinates use the input-image coordinate system and are paired with lane, entry, conflict-zone, and exit descriptions. Scenario-specific constraints may prohibit overtaking, lane changes, identity swaps, or preserve-set motion.

\subsection{Retry and Failure Policy}
Automatic retries are limited to requests that return no media. Incorrect outputs are not resubmitted, and timeouts, policy blocks, malformed files, and incorrect dimensions count as failures. The repair study allows one semantic retry after a Full-interface failure is scored and assigned a localizable dominant cause; both attempts are retained.

\section{Metric Definitions and Detailed Behavior-Reasoning Audit}
\label{app:reasoning}

\subsection{Metric Definitions}
For sample $i$, let $Y_i$ denote the reference behavior-label set and $\widehat{Y}_i$ the predicted set. Sample-set F1 is
\begin{equation}
F_{1,\mathrm{set}}
= \frac{1}{N}\sum_{i=1}^{N}
\frac{2|Y_i\cap\widehat{Y}_i|}{|Y_i|+|\widehat{Y}_i|},
\end{equation}
where an unparsable response contributes zero under the fixed schema. Exact-set match is
\begin{equation}
\mathrm{Exact}
= \frac{1}{N}\sum_{i=1}^{N}\mathbf{1}[Y_i=\widehat{Y}_i],
\end{equation}
and macro label F1 is the unweighted mean of the nine label-wise F1 scores. Any overlap records whether $Y_i\cap\widehat{Y}_i\neq\emptyset$. Parse success is the fraction of responses that conform to the required output schema without manual correction.

Table~\ref{tab:metric-inventory} defines the remaining stage-specific metrics. Continuous diagnostics accompany the binary decisions but do not alter them.
The first five metrics concern image editing, the next five concern video generation, and the final two concern paired factual and counterfactual evaluation.

\begin{center}
\begin{minipage}{\textwidth}
\centering
\small
\begin{tabular}{P{0.23\textwidth}P{0.71\textwidth}}
\toprule
Metric & Operational Definition \\
\midrule
Removal completion & The selected actor is absent from the edited anchor without a visible duplicate or residual instance. \\
Insertion completion & The requested actor is present in the designated region and visually integrated with the scene. \\
Lane-compatible geometry & The inserted actor has a location, heading, scale, perspective, and road contact consistent with the assigned lane. \\
Non-target preservation & Unedited actors and scene regions remain consistent with the input, except for local changes required by the intervention. \\
Outside-mask LPIPS & LPIPS is computed outside the edit mask after spatial alignment; lower values indicate less unintended change. \\
Identity persistence & The bound controlled actor remains identifiable throughout the required visible interval. \\
Route success & The controlled actor remains within the legal corridor and reaches the programmed endpoint when required. \\
Interaction validity & The programmed yield or passage order is satisfied in the shared conflict zone. \\
Preservation validity & Preserve-set actors, the background, and the actor count remain consistent with the scenario. \\
Hallucination rate & The fraction of outputs containing a persistent unrequested actor outside an allowed entry region. \\
Pair validity & Both factual and counterfactual branches satisfy all four binary predicates. \\
Intervention-response consistency (IRC) & Among valid pairs, the affected actor changes timing or passage order in the direction specified by the intervention. \\
\bottomrule
\end{tabular}
\normalsize
\captionof{table}{Stage-specific metric definitions.}
\label{tab:metric-inventory}
\end{minipage}
\end{center}
Table~\ref{tab:metric-inventory} defines the secondary metrics used
throughout the benchmark. Aggregate image-editing results are
reported in Table~\ref{tab:image-editing}, generator validity and
hallucination rates are reported in the main tables, and paired
response results are reported in Table~\ref{tab:paired-response}.

Edit-preservation and perceptual diagnostics that are not aggregated in the main tables remain available in the per-output evaluation
records. They are used for diagnosis and do not override a failed traffic predicate.

\subsection{Aggregate and Class-Conditional Reasoning Audit}
Tables~\ref{tab:reasoning-audit} and~\ref{tab:reasoning-classes} report Gemini 2.5 Pro on the 1,500-sample audit. Support counts positive labels in this audit. Confidence intervals use clip-level bootstrap, with samples from each clip kept together; parse success has no bootstrap interval.

\begin{center}
\begin{minipage}{\textwidth}
\centering
\small
\begin{tabular}{lcc}
\toprule
Metric & Estimate (\%) & 95\% CI \\
\midrule
Parse success & \planned{\pReasonParse} & \planned{\pReasonParseCI} \\
Sample-set F1 & \planned{\pGeminiProSample} & \planned{\pReasonSampleCI} \\
Any overlap & \planned{\pReasonOverlap} & \planned{\pReasonOverlapCI} \\
Exact-set match & \planned{\pReasonExact} & \planned{\pReasonExactCI} \\
Macro label F1 & \planned{\pGeminiProMacro} & \planned{\pReasonMacroCI} \\
\bottomrule
\end{tabular}
\normalsize
\captionof{table}{Aggregate behavior-reasoning audit over 1,500 samples. Percentages use one decimal place.}
\label{tab:reasoning-audit}
\end{minipage}
\end{center}

\begin{center}
\begin{minipage}{\textwidth}
\centering
\small
\maxsizebox{\textwidth}{!}{%
\begin{tabular}{llrrrr}
\toprule
Family & Event Label & Support & Precision (\%) & Recall (\%) & F1 (\%) \\
\midrule
Longitudinal & stationary & \planned{\pStationarySupport} & \planned{\pStationaryPrecision} & \planned{\pStationaryRecall} & \planned{\pStationaryFOne} \\
Longitudinal & keep moving & \planned{\pKeepSupport} & \planned{\pKeepPrecision} & \planned{\pKeepRecall} & \planned{\pKeepFOne} \\
Route-scale & straight trend & \planned{\pStraightSupport} & \planned{\pStraightPrecision} & \planned{\pStraightRecall} & \planned{\pStraightFOne} \\
Longitudinal & slowing or stop & \planned{\pSlowSupport} & \planned{\pSlowPrecision} & \planned{\pSlowRecall} & \planned{\pSlowFOne} \\
Route-scale & left turn & \planned{\pLeftTurnSupport} & \planned{\pLeftTurnPrecision} & \planned{\pLeftTurnRecall} & \planned{\pLeftTurnFOne} \\
Route-scale & right turn & \planned{\pRightTurnSupport} & \planned{\pRightTurnPrecision} & \planned{\pRightTurnRecall} & \planned{\pRightTurnFOne} \\
Lateral & left shift & \planned{\pLeftShiftSupport} & \planned{\pLeftShiftPrecision} & \planned{\pLeftShiftRecall} & \planned{\pLeftShiftFOne} \\
Lateral & right shift & \planned{\pRightShiftSupport} & \planned{\pRightShiftPrecision} & \planned{\pRightShiftRecall} & \planned{\pRightShiftFOne} \\
Longitudinal & start moving & \planned{\pStartSupport} & \planned{\pStartPrecision} & \planned{\pStartRecall} & \planned{\pStartFOne} \\
\bottomrule
\end{tabular}}
\normalsize
\captionof{table}{Class-conditional behavior recognition. Support is the number of positive labels in the 1,500-sample audit.}
\label{tab:reasoning-classes}
\end{minipage}
\end{center}

The 25 dual-state windows belong to the full annotation set, whereas the 13 exclusions are audit samples within the ambiguity margin. Their route-scale labels are retained, but they are omitted from longitudinal scoring, leaving 1,487 records.

\section{Component, Oracle, and Paired Results}
\label{app:component-results}

\subsection{Image-Editing Components}

Table~\ref{tab:image-editing} reports the image-editing results.

\begin{center}
\begin{minipage}{\textwidth}
\centering
\small
\maxsizebox{\textwidth}{!}{%
\begin{tabular}{lrrrr}
\toprule
Editor & Removal $\uparrow$ & Insertion $\uparrow$ & Geometry $\uparrow$ & LPIPS $\downarrow$ \\
\midrule
GPT Image 2 & \planned{\pImageTwoRemoval} & \planned{\pImageTwoInsertion} & \planned{\pImageTwoGeometry} & \planned{\pImageTwoLPIPS} \\
Gemini 3 Pro Image & \planned{\pGeminiImageRemoval} & \planned{\pGeminiImageInsertion} & \planned{\pGeminiImageGeometry} & \planned{\pGeminiImageLPIPS} \\
Qwen-Image-2.0 Pro & \planned{\pQwenImageRemoval} & \planned{\pQwenImageInsertion} & \planned{\pQwenImageGeometry} & \planned{\pQwenImageLPIPS} \\
Seedream 5.0 Pro & \planned{\pSeedreamRemoval} & \planned{\pSeedreamInsertion} & \planned{\pSeedreamGeometry} & \planned{\pSeedreamLPIPS} \\
\bottomrule
\end{tabular}}
\normalsize
\captionof{table}{Image-editing results for 60 cases, with 30 removal and 30 insertion cases. Each case is repeated twice. Removal and insertion are evaluated on 60 outputs per editor, geometry on 60 insertion outputs, and LPIPS outside the edit mask.}
\label{tab:image-editing}
\end{minipage}
\end{center}

Outside-mask LPIPS measures unintended change outside the edited region and is reported separately from completion, geometry, and non-target preservation.

\subsection{Oracle Substitutions}

Table~\ref{tab:oracle} reports the matched oracle substitutions.

\begin{center}
\begin{minipage}{\textwidth}
\centering
\small
\begin{tabular}{lr}
\toprule
Input Setting & E2E (\%) \\
\midrule
Predicted program + edited anchor & \planned{\pFullEnd} \\
Audited program + edited anchor & \planned{\pOracleProgramEnd} \\
Predicted program + audited anchor & \planned{\pOracleAnchorEnd} \\
Audited program + audited anchor & \planned{\pOracleBothEnd} \\
\bottomrule
\end{tabular}
\normalsize
\captionof{table}{Oracle substitutions for a fixed generator across 60 matched scenario and repeat slots per input setting. The four settings yield 240 records. Values are end-to-end success percentages.}
\label{tab:oracle}
\end{minipage}
\end{center}

In each oracle setting, only the named upstream component is replaced; the scenario, generator, and repeat slot remain fixed. The gains therefore need not be additive.

\subsection{Paired Factual and Counterfactual Response}

Table~\ref{tab:paired-response} reports the paired factual and
counterfactual results.

\begin{center}
\begin{minipage}{\textwidth}
\centering
\small
\maxsizebox{\textwidth}{!}{%
\begin{tabular}{lrrrrr}
\toprule
Intervention & Valid Pair (\%) & IRC (\%) & $\Delta$ Entry & $\Delta$ Wait & Order Change (\%) \\
\midrule
Removal ($n=24$) & \planned{\pRemovalPairValid} & \planned{\pRemovalResponse} & \planned{\pRemovalEntryDelta} & \planned{\pRemovalWaitDelta} & \planned{\pRemovalOrderChange} \\
Insertion ($n=21$) & \planned{\pInsertionPairValid} & \planned{\pInsertionResponse} & \planned{\pInsertionEntryDelta} & \planned{\pInsertionWaitDelta} & \planned{\pInsertionOrderChange} \\
Overall ($n=45$) & \planned{\pOverallPairValid} & \planned{\pOverallResponse} & -- & -- & -- \\
\bottomrule
\end{tabular}}
\normalsize
\captionof{table}{Paired factual and counterfactual response. Pair validity uses all matched pairs. IRC and order change are computed only for valid pairs. The reported time differences are medians over valid pairs, and $\Delta$ is counterfactual minus factual time in seconds.}
\label{tab:paired-response}
\end{minipage}
\end{center}

Removal is expected to advance entry, reduce waiting, or move the affected actor earlier in the passage order; insertion is expected to have the opposite effect. All $\Delta$ values are counterfactual minus factual.

\section{Failure Taxonomy and One-Step Repair}
\label{app:failure}

Table~\ref{tab:failure-taxonomy} lists the repair-oriented failure
categories and their permitted one-step responses.

\begin{center}
\begin{minipage}{\textwidth}
\centering
\small
\maxsizebox{\textwidth}{!}{%
\begin{tabular}{P{0.16\textwidth}P{0.13\textwidth}P{0.37\textwidth}P{0.27\textwidth}}
\toprule
Failure & Predicate & Detection Signal & Permitted One-Step Response \\
\midrule
Actor reappearance or duplication & Initial / preservation & A removed identity returns, or two tracks match the same bound target. & Add persistent-absence and actor-count invariants. \\
Illegal exit or route drift & Behavioral & The target footpoint leaves the legal corridor or terminates at the wrong exit. & Narrow the corridor and expose one additional route anchor. \\
Scale or heading mismatch & Initial & Inserted box geometry or orientation conflicts with the assigned lane and scene depth. & Restate the lane-aligned heading, scale, and road-contact constraints. \\
Non-target motion & Preservation & A preserve-set track changes state beyond the behavior permitted by the scenario without a corresponding interaction path. & Strengthen the actor-specific stationary or motion invariant. \\
Hallucinated actor & Preservation / relational & A new unmatched track persists outside an allowed entry region. & Add a no-new-actor rule and an explicit actor-count bound. \\
Coupled failure & Multiple & Two or more dependent predicate violations occur in sequence. & Record as coupled; no single-predicate repair is credited. \\
\bottomrule
\end{tabular}}
\normalsize
\captionof{table}{Repair-oriented failure taxonomy used by the verifier. Localized categories have predefined one-step responses; coupled failures are recorded but are not repaired. The table is not a frequency distribution.}
\label{tab:failure-taxonomy}
\end{minipage}
\end{center}

Each failure receives one dominant label, with secondary labels retained. One localizable failure may trigger one corrective condition and one retry; coupled failures are not repaired. The Full~+~One Repair result therefore comprises 60 final decisions and 27 additional calls.

\section{Scenario Manifest and Release Records}
\label{app:scenario}

Scenario, execution, score, and relation records are stored separately. Actor programs are stored independently of prompt text, and Table~\ref{tab:manifest-records} lists the retained fields.

\begin{center}
\begin{minipage}{\textwidth}
\centering
\small
\maxsizebox{\textwidth}{!}{%
\begin{tabular}{P{0.17\textwidth}P{0.76\textwidth}}
\toprule
Record & Required Fields \\
\midrule
Scenario record & Scenario identifier; source clip and frame range; site and illumination; intervention type; target, interacting, and preserve-set actors; topology version; input-image checksum; actor program; legal route; temporal window. \\
Insertion record & Entry lane; initial box; heading; requested scale range; edited-anchor identifier. \\
Removal record & Original target track; region required to remain empty; persistence requirement. \\
Execution record & Model and provider identifier; access timestamp; input resolution and reference order; requested duration; seed when exposed; prompt hash; transport retry count; latency; returned status, dimensions, and error code. \\
Score record & Four binary predicates; continuous diagnostics; dominant and secondary failure labels; independent rater decisions; adjudicated decision; scoring-software version. \\
Relation record & Factual partner and repeat index for paired tests; substituted component for oracle tests; first-pass parent and corrective rule for repair tests. \\
\bottomrule
\end{tabular}}
\normalsize
\captionof{table}{Logical records retained for benchmark traceability.}
\label{tab:manifest-records}
\end{minipage}
\end{center}

Failed calls remain linked to their scenarios, and retries do not overwrite first-pass records. Annotations, topology, scenarios, and adapters use separate version fields. Paths are relative, identifying metadata are removed, and available media checksums are stored with the records.

\section{Additional Counterfactual Example}
\label{app:additional-example}

Table~\ref{tab:additional-scenario} summarizes the insertion scenario used for the video rollouts in Figure~\ref{fig:additional-video}. The case is excluded from all aggregate results.

\begin{center}
\begin{minipage}{\textwidth}
\centering
\small
\begin{tabular}{ll}
\toprule
Field & Value \\
\midrule
View and illumination & Second fixed-camera viewpoint; daytime \\
Intervention & Insert a dark-blue sedan T1 \\
Interacting actor & White SUV T2 ahead of T1 \\
Requested route & Same straight image-plane lane toward the upper-left exit \\
Interaction order & T2 moves first; T1 starts approximately 0.3--0.5 s later \\
Temporal horizon & 8 s \\
Preserve set & School bus and all unboxed road users \\
Shared video anchor & GPT Image 2 insertion output \\
\bottomrule
\end{tabular}
\normalsize
\captionof{table}{Specification of the insertion scenario used for the
additional video rollouts.}
\label{tab:additional-scenario}
\end{minipage}
\end{center}

\begin{center}
\begin{minipage}{\textwidth}
    \centering
    \includegraphics[width=\textwidth]{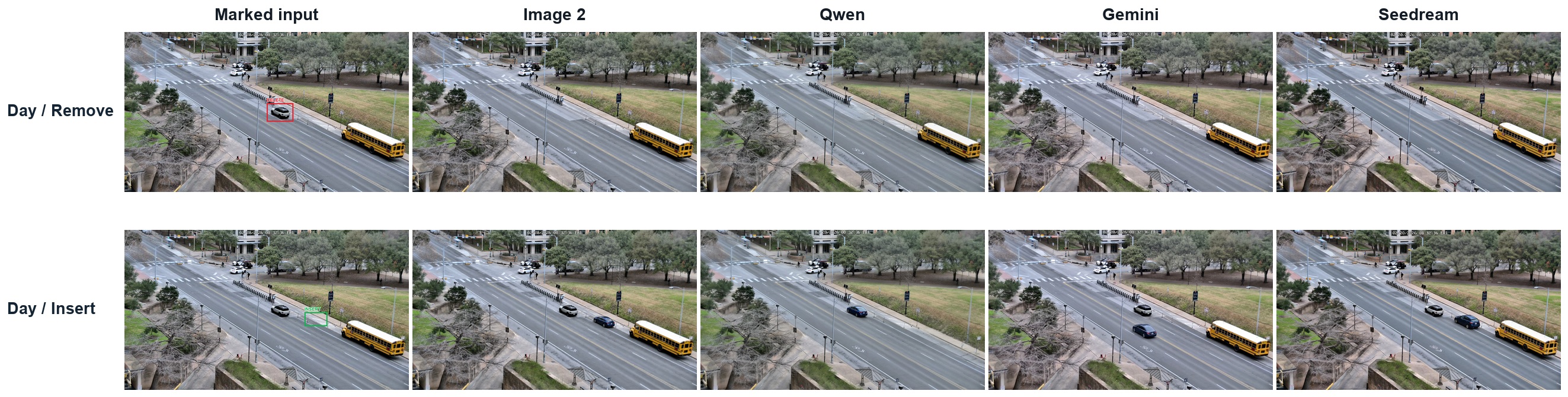}
    \captionof{figure}{Additional removal and insertion anchors from the second fixed-camera viewpoint. The marked input is followed by outputs from GPT Image 2, Qwen-Image-2.0 Pro, Gemini 3 Pro Image, and Seedream 5.0 Pro.}
    \label{fig:additional-editing}
\end{minipage}
\end{center}

Figure~\ref{fig:additional-editing} shows the removal and insertion
anchors. Removal is evaluated for road reconstruction and
preservation. Insertion also requires lane-consistent heading,
scale, illumination, occlusion, and road contact. The GPT Image 2
insertion is shared across all rollouts in
Figure~\ref{fig:additional-video}.

\begin{center}
\begin{minipage}{\textwidth}
    \centering
    \includegraphics[width=\textwidth]{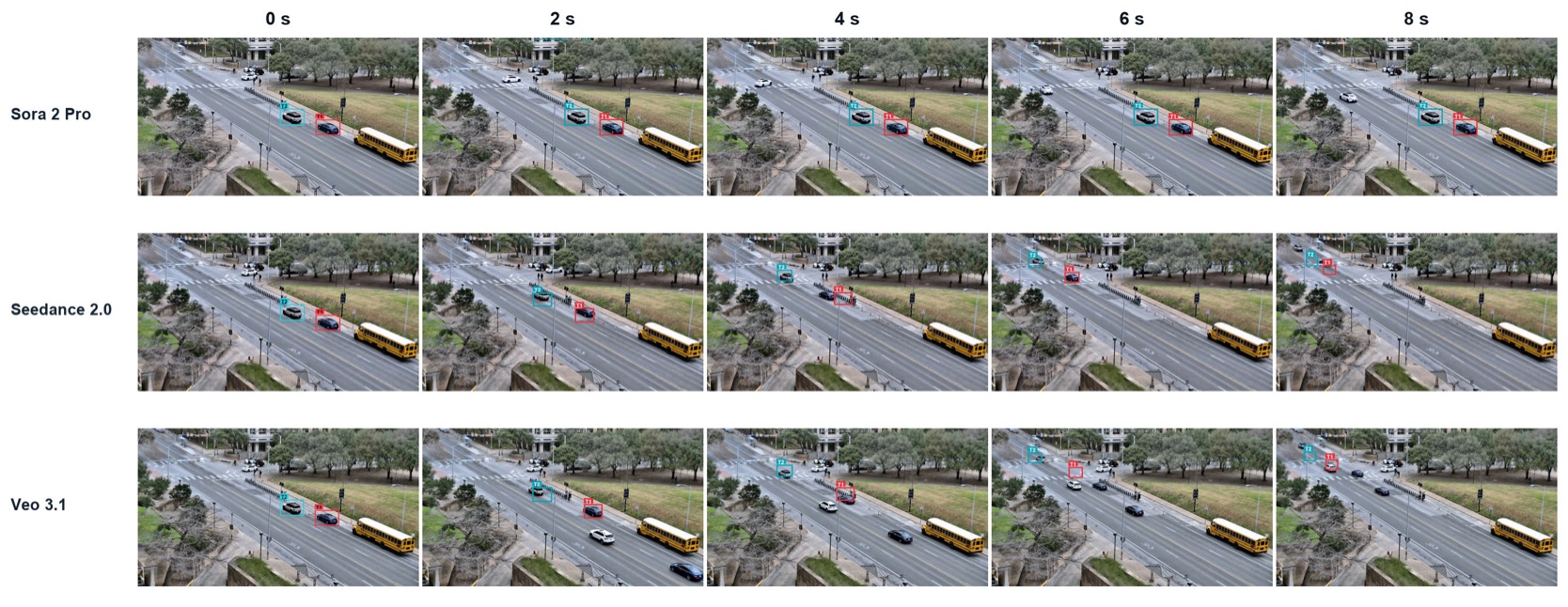}
    \captionof{figure}{Eight-second rollouts for the insertion case. T1 is the inserted dark-blue sedan and T2 the leading SUV. Both are instructed to follow the same image-plane lane toward the upper-left exit, with T2 moving first and T1 maintaining a visible non-overlapping gap. Red and cyan boxes mark T1 and T2.}
    \label{fig:additional-video}
\end{minipage}
\end{center}

All rollouts use the settings in Table~\ref{tab:generator-config}, including \texttt{veo-3.1-generate-001} at 1080p for Veo. The supplied image defines time zero; T2 starts first, T1 starts 0.3--0.5~s later, and anchors are specified every 2~s. The camera, road, school bus, and unboxed vehicles are required to remain fixed, and T1 is instructed not to overtake or overlap T2.

In this case, Seedance follows the requested order and route. Sora leaves both vehicles nearly stationary. Veo moves them in the requested direction but introduces additional vehicles, failing preservation.
\end{document}